\documentclass[a4paper, fleqn]{cas-sc}
\usepackage{graphicx}%
\usepackage{multirow}%
\usepackage{amsmath,amssymb,amsfonts}%
\usepackage{amsthm}%
\usepackage{mathrsfs}%
\usepackage{adjustbox}
\usepackage[title]{appendix}%
\usepackage{xcolor}%
\usepackage{textcomp}%
\usepackage{manyfoot}%
\usepackage{booktabs}%
\usepackage{algorithm}%
\usepackage{makecell}
\usepackage{caption}
\usepackage{algorithmicx}%
\usepackage{algpseudocode}%
\usepackage{listings}%
\usepackage{tabularx}
\usepackage{array}
\usepackage{booktabs}
\usepackage{rotating}
\usepackage{multirow}
\usepackage{tikz}
\usepackage[numbers]{natbib}
\usetikzlibrary{shapes.geometric, arrows.meta, positioning}
\PassOptionsToPackage{hidelinks}{hyperref}
\usepackage{hyperref}

\theoremstyle{thmstyleone}%
\newcolumntype{C}{>{\centering\arraybackslash}X}
\newcolumntype{P}[1]{>{\centering\arraybackslash}p{#1}} 
\theoremstyle{thmstyletwo}%
\theoremstyle{thmstylethree}%

\begin{document}
\let\WriteBookmarks\relax
\def\floatpagepagefraction{1}
\def\textpagefraction{.001}
\shorttitle{Leveraging social media news}
 \shortauthors{Xing Hu et~al.}

\title [mode = title]{A Comprehensive Review of Diffusion Models in Smart Agriculture: Progress, Applications, and Challenges
\renewcommand{\thefootnote}{}%
\footnote{The preprint version of this paper appears in \url{https://arxiv.org/abs/2507.18376}}%
\addtocounter{footnote}{-1}%
\renewcommand{\thefootnote}{\arabic{footnote}}%
}                   

\author[1]{Xing Hu}
\ead{huxing@usst.edu.cn}
\affiliation[1]{
    organization={School of Optical-Electrical and Computer Engineering, University of Shanghai for Science and Technology},
    addressline={No. 516, Jungong Road},
    city={Shanghai},
    postcode={200093},
    country={China}
}

\author[1]{Haodong Chen}
 \ead{18856715535@163.com}

\author[2]{Qianqian Duan}
 \ead{dqq1019@163.com}
\affiliation[2]{
    organization={School of Electronics and Electrical Engineering, Shanghai University of Engineering Science},
    addressline={No. 333, Longteng Road},
    city={Shanghai},
    postcode={201620},
    country={China}
}

\author[1]{Dawei Zhang}
\cormark[1]
\ead[cor1]{dwzhang@usst.edu.cn}

\cortext[cor1]{Corresponding author}



\begin{abstract}
The global population is rising, while arable land resources are becoming increasingly scarce. Smart and precision agriculture are therefore essential for sustainable development. Artificial intelligence (AI), especially deep learning, has been widely used in crop monitoring, pest detection, and yield prediction. Among recent generative models, diffusion models have shown strong potential in agricultural image processing, data augmentation, and remote sensing. Compared with traditional generative adversarial networks (GANs), they provide more stable training and higher image quality.They also mitigate challenges such as limited labeled data and class imbalance. In recent years, many studies have applied diffusion models to agriculture. However, there is still no detailed survey that summarizes, evaluates, and categorizes these methods. This paper presents the first comprehensive review of diffusion models in agriculture. We focus on crop disease and pest detection, remote sensing image enhancement, crop growth prediction, and agricultural resource management. Studies show that diffusion models can improve the accuracy, robustness, and generalization of downstream models. They are effective for image synthesis, augmentation, and denoising in complex environments. Challenges remain in computational cost and cross-domain adaptation. However, diffusion models are expected to play a growing role in intelligent agriculture. As the technology develops, it holds promise for improving food security and supporting environmental sustainability.
\end{abstract}

\maketitle

\section{Introduction}\label{sec1}

Traditional agriculture is facing unprecedented challenges as the global population rises and the amount of arable land decreases. To improve efficiency, ensure food security, and achieve sustainable development, nations worldwide are pushing forward with agricultural modernization. Within this context, smart agriculture and precision farming are gradually emerging as central pillars for the future of farming, playing a crucial role in the overall modernization process.

Smart agriculture brings together advanced information technologies—including artificial intelligence (AI), the Internet of Things (IoT), big data, and remote sensing—to enable real-time monitoring and data-driven decision-making for factors such as crop growth, equipment status, soil moisture, and climate variability. Through approaches like precise fertilization, targeted irrigation, and intelligent pest and disease control, resources can be used more effectively, while making agricultural practices smarter, more automated, and more sustainable. With the rapid progress of AI, deep learning models for image generation have become a research hotspot in agricultural intelligence. Among them, diffusion models—an emerging class of generative models—have shown remarkable performance in tasks such as image generation~\cite{1,2,3,4,5,6,29}, restoration~\cite{1,3,4,7,8,9}, segmentation~\cite{10,11,12,13}, and enhancement~\cite{28}. They have already achieved notable success in fields including medicine~\cite{10,12}, industrial inspection~\cite{14}, and digital art. Applying these models in agricultural contexts offers promising solutions to persistent challenges like limited datasets, sample imbalance, and insufficient data diversity.

This paper aims to systematically review the current state and typical applications of diffusion models in agriculture, thereby promoting the integration of AI and modern farming, and accelerating the practical implementation of Agricultural intelligence. Deep learning, which is known for its ability to find patterns and extract useful features, has become an important technology for automating and informatizing agriculture. Deep learning is different from regular machine learning because it can learn complicated nonlinear relationships on its own, from start to finish. This facilitates the analysis of images, time series, and multisource data. It has been applied in several fields. Examples include crop identification~\cite{15,16}, pest and disease detection~\cite{15,16}, yield prediction~\cite{17}, and environmental monitoring~\cite{19,23,25}.

Convolutional neural networks (CNNs)~\cite{30} are extensively utilized in image processing. They are used for crop classification~\cite{31}, weed detection~\cite{31}, and leaf disease diagnosis~\cite{15,16}. High-resolution images help these models detect small differences in field conditions. In remote sensing, deep neural networks handle radar~\cite{20,22} and multispectral data~\cite{21,24,26} to identify crop types, planting areas, growth stages, and water content. For sequence data, recurrent neural networks (RNNs) and long short-term memory (LSTM)~\cite{25} are used to model crop growth, soil moisture, and weather factors. These models help make better decisions about things like watering and fertilizing.

Generative models have also been explored. Generative adversarial networks (GANs)~\cite{28,32} are used to create synthetic data and balance sample distributions. Yet, their training often suffers from instability and mode collapse, limiting practical use. Diffusion models avoid some of these issues by providing more stable learning and producing higher-quality outputs. As a result, they are gaining importance in agricultural applications.

 Significant advances in generative modeling have recently been made by the class of deep generative models known as diffusion models.  The mathematical simulation of diffusion processes from non-equilibrium thermodynamics serves as their basis.  In contrast to VAEs ~\cite{33,34} and GANs~\cite{28,32}, diffusion models generate high-quality outputs in two stages.  Noise is progressively introduced into the data during training (forward process).  After that, the model learns to reverse this process by gradually denoising (reverse process).  Ho et al. introduced the denoising diffusion probabilistic model (DDPM) in 2020. It was the first large-scale use of deep learning and helped make it more stable.  Later, DDIM speeded up generation by using a reverse process that wasn't Markovian.  Score-based generative models (SGMs) improved the field by learning score functions to overcome some limits of earlier methods. Conditional diffusion models allowed flexible tasks such as text-to-image generation. Latent diffusion models (LDMs) were introduced in 2022. They operate in latent space, lower computational cost, and support high-resolution generation. Stable Diffusion further advanced the area by providing open-source tools for text-to-image tasks.

Despite these strengths, diffusion models still have drawbacks. They need large computation, depend on data quality, and often lack generalization. Future studies are expected to improve efficiency, support multimodal fusion, and adapt to low-resource settings. These advances will enhance the role of generative AI in agriculture and support its wider adoption.

\begin{figure}[htbp]
\centering
\includegraphics[width=0.7\textwidth]{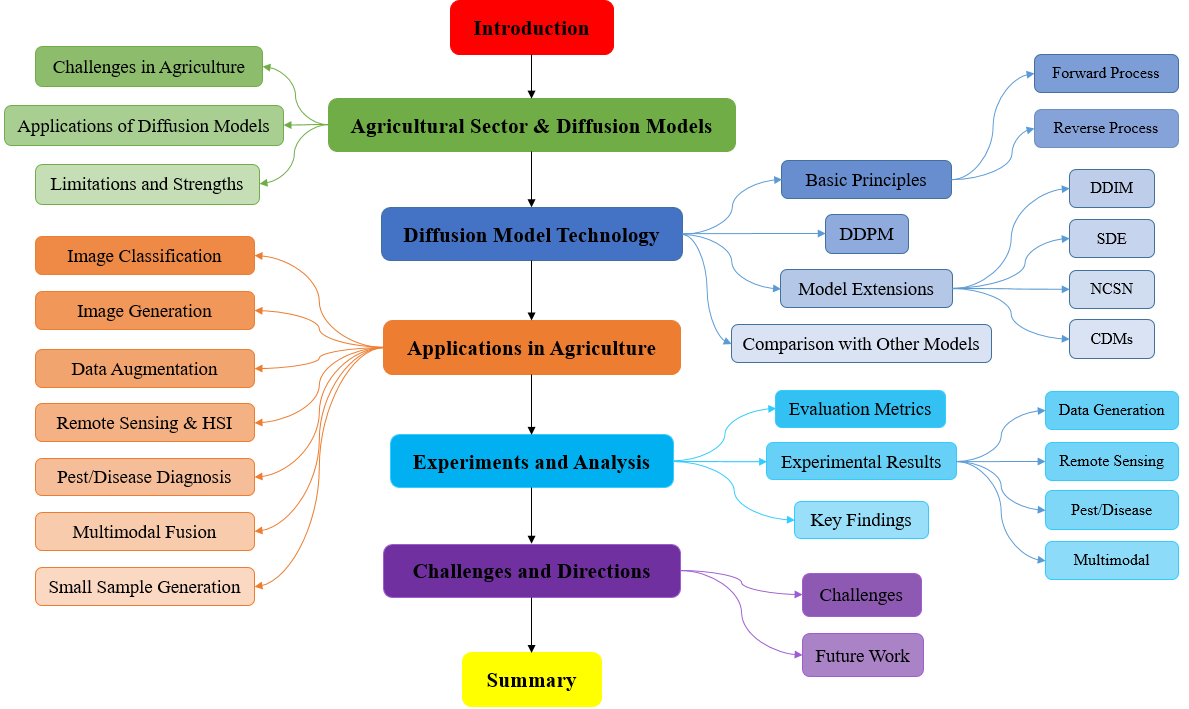}%
\caption{Structural overview of this review on diffusion models in agriculture.}
\label{fig0}
\end{figure}

The key points of this paper are as follows:

\begin{enumerate}
    \item \textbf{Survey and Framework:} We present a review of diffusion models in smart agriculture. Their uses are grouped into seven areas: image classification, image generation, data augmentation, remote sensing and hyperspectral reconstruction, pest and disease detection, multimodal fusion, and small-sample data generation. Figure~\ref{fig0} shows this framework in a clear and simple way.

    \item \textbf{Empirical Benchmarking and Practical Validation:} In contrast to previous theoretical surveys, we perform small-scale empirical studies to evaluate diffusion models (such as DDPM and Stable Diffusion) against conventional GANs in standard agricultural tasks. We specifically confirm that diffusion models work well for solving long-tail distribution problems by using synthetic data augmentation. Our results show that the proposed method is more accurate, stable, and generalizable in challenging agricultural scenarios.

\item \textbf{Analysis of Model Variants:} We take a closer look at the main types of diffusion models, including DDIM, Score-based Models, Conditional Models, and Latent Models. Their theory, design, and use in agriculture are described. We also discuss their strengths and practical value.

\item \textbf{Challenges and Future Work:} Right now, these models take a lot of computing power, need plenty of data, and don’t always work well in new situations.
 Future research should look into lightweight models, better use of multimodal data, domain adaptation methods, and real-time applications to make diffusion models more practical in agriculture.  

\item \textbf{AI and Agriculture Integration:} This study not only serves as a technical reference but also shows how diffusion models can help improve food security, support sustainable farming, and strengthen environmental monitoring. It also points out future directions to bring artificial intelligence and agriculture closer together.

\end{enumerate}

The structure of this paper is arranged as follows:  
Section~\ref{sec1} sets out the motivation and background, and explains why diffusion models are becoming important in modern agriculture.  
Section~\ref{sec2} looks at the types of data available in agriculture, the technical requirements, and the practical challenges that need to be addressed.  
Section~\ref{sec3} reviews the basic theory, the development process, and the main categories of diffusion models, with attention to recent progress in generative modeling.  
Section~\ref{sec4} turns to practical applications in agricultural image analysis, covering tasks such as classification, segmentation, data augmentation, image generation, and pest and disease detection.  
Section~\ref{sec5} shows the experimental results and compares the performance of the models on several agricultural datasets.  
Section~\ref{sec6} looks at the main limitations and open questions, and puts forward some possible solutions and ideas for future research.  
Finally, Section~\ref{sec7} wraps up the paper by summarizing the main findings and contributions, and by talking about how diffusion models could influence the future of smart and precision agriculture.

\section{An Overview of the Agricultural Sector and the Context for Using the Diffusion Model}\label{sec2}

\subsection{Key Problems and Technology Needs for Farming}

There are a number of real-world problems that modern farming has to deal with. These include keeping crops healthy, managing pests and diseases, using water and fertilizer efficiently, and making sense of the huge amount of data from remote sensing and sensors.
These include managing the massive volumes of data generated by remote sensing technologies, controlling pests and diseases, and making better use of land, water, and fertilizers.
Pests and diseases remain one of the biggest threats to yields. Traditional manual checks are often slow and unreliable, especially when problems are just beginning to appear. As farms grow larger, it has become harder to rely only on human inspection. New technologies like computer vision and artificial intelligence make it possible for sensors and remote sensing tools to keep an eye on the health of crops in real time. This means you won't have to check things by hand as often, and it will be easier to find pests and diseases early on.  

The goal of precision agriculture is to use resources more efficiently, raising yields while reducing environmental impact. Farmers can adjust planting time, water, fertilizer, and pesticide use as needed. This approach helps lower carbon emissions and supports sustainable farming. It also relies on big data, remote sensing, and IoT sensors to monitor weather, soil, and crops, giving farmers reliable information for decision-making. As agriculture becomes more important, remote sensing is playing a bigger role, especially for large-scale monitoring of crop health and farmland conditions.

\begin{figure}[htbp]
\centering
\includegraphics[width=0.7\textwidth]{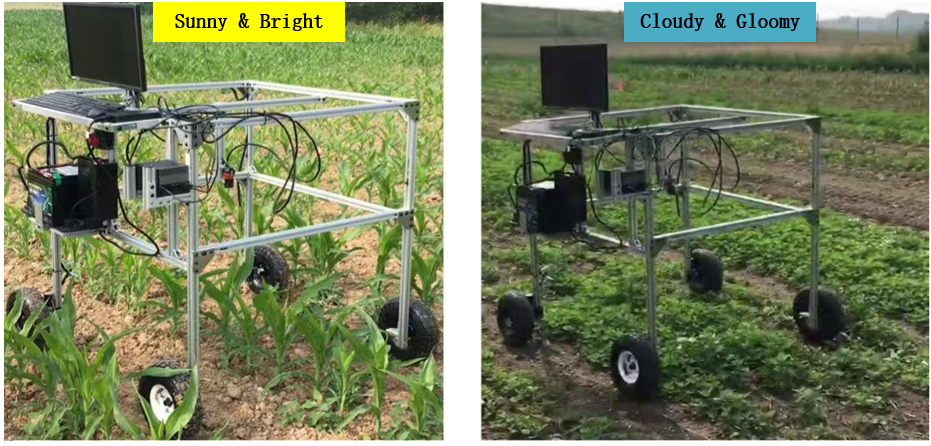}%
\caption{Weed image acquisition by a mobile platform with OpenWeedGUI in two different fields~\cite{146}.}
\label{image1}
\end{figure}

As illustrated in Figure~\ref{image1}, mobile platforms can be used to collect weed imagery. Despite the advantages of real-time data acquisition, remote sensing images are often affected by weather and lighting conditions, leading to degraded quality. Improving the resolution, clarity, and noise reduction of these images is still a major challenge. In addition, handling and analyzing large remote sensing datasets efficiently to get useful agricultural information is also an urgent problem.

To address these challenges, there is a growing need to integrate intelligent and automated technologies, such as deep learning-based pest and disease detection systems, decision support systems for precision agriculture, and advanced image enhancement techniques for remote sensing. These innovations offer the potential to significantly improve agricultural productivity, decision-making accuracy, and sustainability, facilitating the transition from traditional farming to intelligent, precision-driven agriculture.


\subsection{Potential Applications of Diffusion Models in Agriculture}

In recent years, diffusion models have gradually attracted attention in agricultural research, and are mainly used in tasks such as crop monitoring, pest and disease identification, and remote sensing image analysis. In the context of limited labeled data or uneven distribution of categories, the diffusion model can generate synthetic samples to assist training, thereby improving the performance of downstream tasks.

Taking the early identification of crop diseases and insect pests as an example, many regions still rely on manual inspection, but this method is time-consuming and laborious, and it is often difficult to detect early subtle symptoms in time. The diffusion model can generate high-resolution images of lesions or insect pests, which can be used to expand the data set and improve the effect of the automatic detection system.

\begin{figure}[htbp] \centering \includegraphics[width=0.7\textwidth]{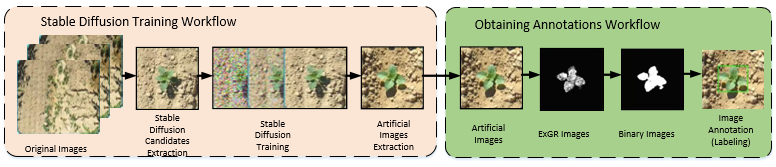} \caption{Artificial image generation process based on Stable Diffusion training~\cite{145}.} \label{image2} \end{figure}

By incorporating variables such as soil moisture and climate change, diffusion models can generate simulated crop growth data under diverse environmental conditions (Figure~\ref{image2}), enabling more accurate predictions of crop health and yield and offering scientific support for planting and management decisions.

In addition, diffusion models have shown great potential in remote sensing image enhancement and denoising. Although remote sensing can provide valuable information for agricultural monitoring, the image quality is often disturbed by environmental noise. Diffusion-based enhancement methods can effectively improve the clarity and resolution of images, and this improvement helps to more accurately carry out land use assessment, crop growth monitoring and climate impact analysis.

\subsection{Advantages and Challenges of Diffusion Models in Agriculture}

The advantages of diffusion models in agricultural applications are mainly reflected in high-quality image generation, strong stability of the training process, and good support for conditional generation. Compared with the traditional generative adversarial network (GAN), the diffusion model is more stable in training and is not prone to problems such as mode collapse, so it can generate more diverse and realistic agricultural images.

First, diffusion models achieve high-quality image generation through progressive denoising, which is particularly useful in complex agricultural scenarios. The model can create detailed and realistic images for tasks like pest detection or crop growth simulation, making the results closer to real agricultural conditions and improving the reliability and clarity of data-driven decisions.
Unstable convergence and mode collapse are typical problems that generative adversarial networks (GANs) often face in training. In contrast, diffusion models rely on a progressive denoising process that ensures more stable, robust, and reproducible training. This feature makes it suitable for large-scale agricultural image generation tasks.

Third, diffusion models have the ability to generate images conditionally, based on crop type, pest type, or environmental conditions. This flexibility helps to develop customizable agricultural intelligence solutions, thereby supporting more accurate monitoring, decision-making, and resource optimization.

Nevertheless, diffusion models still face some challenges. The large computational overhead, slow inference speed, and dependence on high-quality training data limit their large-scale application in actual agricultural scenarios. Future research should focus more on three aspects: first, improving the computational efficiency of the model; second, continuously expanding agricultural-related data sets; and third, actively exploring multi-modal data fusion methods. through these efforts, the practical application of diffusion models in agriculture can be further promoted.

\section{Overview of Diffusion Model Technology}\label{sec3} Diffusion Models (DMs) are a type of deep generative model, and their theoretical basis comes from the principles of probabilistic graphical models.
 They simulate the random perturbation and denoising process of data~\cite{38}, and gradually generate high-quality samples from pure noise. The core idea is to gradually transform data into noise through the forward diffusion process, and then reconstruct the data by using the learned reverse process. Diffusion models originated from the diffusion principle in non-equilibrium thermodynamics, and have achieved remarkable results in image generation~\cite{17,39}, speech synthesis~\cite{43}, medical imaging~\cite{42} and agricultural remote sensing~\cite{40,41}. They excel at processing high-dimensional data, especially in image generation and denoising tasks, and show good application prospects in hyperspectral image analysis~\cite{41}.


\subsection{Basic Principles of Diffusion Models}

\begin{figure}[htbp] \centering \includegraphics[width=0.7\textwidth]{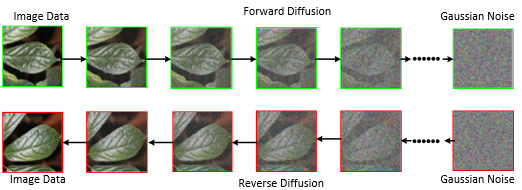} \caption{The forward diffusion and reverse generation process of the diffusion model. \cite{92}} \label{image3} \end{figure}

Diffusion models are built on the basis of Markov processes, and simulate the gradual transformation of data through a noise-driven framework. As shown in Figure~\ref{image3}, the model contains two stages: the first is the forward diffusion process, which gradually destroys the original data by continuously superimposing Gaussian noise; the second is the reverse generation process, which uses the trained neural network to gradually remove the noise, thereby reconstructing realistic data.

\subsubsection{Forward Diffusion Process}

Let the initial sample be $\mathrm{x}_0 \sim q(\mathrm{x}_0)$. During the forward diffusion process, Gaussian noise is superimposed at each step. After $T$ moments, the process is defined as:

\begin{equation} \label{deqn_ex1a1} q(\mathrm{x}_t \mid \mathrm{x}_{t-1}) = \mathcal{N}(\mathrm{x}_t; \sqrt{1 - \beta_t} \, \mathrm{x}_{t-1}, \beta_t I) \end{equation}

where $\beta_t \in (0,1)$ represents the noise variance schedule at time $t$, which can be linear, cosine or learnable.

This process forms a Markov chain. After $t$ steps, the state $\mathrm{x}_t$ can be expressed as a direct sample from the original data $\mathrm{x}_0$:

\begin{equation} 
\label{deqn_ex1a2} 
q(\mathrm{x}_t \mid \mathrm{x}_0) = \mathcal{N}\left(\mathrm{x}_t; \sqrt{\bar{\alpha}_t} \, \mathrm{x}_0, (1 - \bar{\alpha}_t) I\right) 
\end{equation}

Here, $\bar{\alpha}_t$ denotes the cumulative product of the signal preservation factor up to step $t$:

\begin{equation} 
\label{deqn_ex1a3} 
\bar{\alpha}_t = \prod_{s=1}^{t}(1 - \beta_s) 
\end{equation}

With this formulation, $\mathrm{x}_t$ can be derived directly from $\mathrm{x}_0$, eliminating the requirement for sequential sampling across each step, which effectively simplifies the training procedure.

Intuitively, the forward diffusion process adds a small amount of noise to the data at each step. As $t$ increases, the structure and semantics of the data gradually disappear, and finally converge to an isotropic Gaussian distribution.

\subsubsection{Reverse Generation Process}

The reverse generation process is the core part of the diffusion model, which is responsible for gradually transforming pure noise into new samples. The initial input is usually a noise vector sampled from a multivariate Gaussian distribution. The model gradually reconstructs the approximation of the original data through iterative denoising, thereby generating high-quality images or other structured outputs.

In mathematical form, the reverse process is modeled as a parameterized Markov chain that approximates the inverse of the forward noising process:

\begin{equation} \label{deqn_ex1a4} p_\theta(\mathrm{x}_{t-1} \mid \mathrm{x}_t) \end{equation}

Specifically, the model starts from the initial Gaussian noise $\mathrm{x}_T \sim \mathcal{N}(0, I)$ and obtains $\mathrm{x}_{T-1}, \mathrm{x}_{T-2}, \ldots, \mathrm{x}_0$ in turn through a series of denoising steps. The conditional distribution of each step is:

\begin{equation} \label{deqn_ex1a5} p_\theta(\mathrm{x}_{t-1} \mid \mathrm{x}_t) = \mathcal{N}\left(\mathrm{x}_{t-1}; \mu_\theta(\mathrm{x}_t, t), \Sigma_\theta(\mathrm{x}_t, t)\right) \end{equation}

where $\mu_{\theta}$ and $\Sigma_{\theta}$ represent the estimated mean and covariance at time $t$, respectively. These parameters are usually learned through neural networks and used to approximate the inverse operation of the forward noise process.

\subsection{Denoising Diffusion Probabilistic Models (DDPM)}

Modern diffusion generative techniques trace one of their most influential foundations back to 2020, when Ho et al. presented the denoising diffusion probabilistic model (DDPM)~\cite{35,44}, a work that reshaped the trajectory of the field and established a new paradigm for generative modeling.  As a pioneering work, DDPM has greatly promoted the development of generative modeling. In image generation and reconstruction tasks, DDPM can produce high-quality results, and it is superior to earlier generative models such as GANs~\cite{28,32} in terms of stability and theoretical framework.

The core idea of DDPM is to gradually transform clean images into pure noise through a "diffusion process" (implemented by Markov chain), and then use neural networks to learn its inverse process to restore the original images from the noise. The forward process is artificially designed to map any data distribution to a simple prior distribution (such as a standard Gaussian distribution); while the reverse process parameterizes the transfer kernel through a deep neural network, thereby learning the ability to restore the original data.

As mentioned above, DDPM gradually transforms the initial sample $x_0$ into a noisy version $x_t$ through Gaussian perturbation in the forward process. At step $t$, the process can be expressed as:

\begin{equation} \label{deqn_ex1a6} \mathrm{x}_t = \sqrt{\bar{\alpha}_t}\,\mathrm{x}_0 + \sqrt{1 - \bar{\alpha}_t}\,\epsilon,\quad \epsilon \sim \mathcal{N}(0, I) \end{equation}

where $\bar{\alpha}_t = \prod_{s=1}^{t}(1 - \beta_s)$ is used to control the cumulative noise level.

The reverse process is defined by the conditional probability $p_\theta(\mathrm{x}_{t-1} \mid \mathrm{x}_t)$ and parameterized by a neural network. Instead of directly predicting the clean sample, the training objective of DDPM is to estimate the noise $\boldsymbol{\epsilon}$ added in the forward process. The mean estimated by the model at time step $t$ is given by:

\begin{equation} 
\label{deqn_ex1a7} 
\mu_\theta(\mathrm{x}_t, t) = \frac{1}{\sqrt{\alpha_t}}\left(\mathrm{x}_t - \frac{1 - \alpha_t}{\sqrt{1 - \bar{\alpha}_t}}\, \epsilon_\theta(\mathrm{x}_t, t)\right) 
\end{equation}

In the sampling stage, the network repeatedly draws from the corresponding conditional Gaussian distribution, producing a sequence of denoised states that progressively approach the clean data, until the final output $\mathrm{x}_0$ is obtained.

In terms of network structure, DDPM adopts the U-Net architecture~\cite{45} as the denoising backbone network. U-Net was first applied to medical image segmentation, and it adopts an encoder-decoder structure and is equipped with skip connections. The encoder gradually extracts hierarchical features and reduces spatial resolution, while the decoder is responsible for restoring high-resolution output. Skip connections preserve local spatial details, which improves the accuracy of denoising and the quality of generation. In addition, DDPM also introduces time step embedding (usually sinusoidal position encoding) to embed time information into the network, so that it can adaptively denoise for different noise levels.

In terms of training objectives, DDPM minimizes the variational upper bound of the negative log-likelihood by introducing KL divergence, and its formula is:

\begin{equation} \label{deqn_ex1a8} \begin{aligned} \mathbb{E}_{q}\left[-\log p_{\theta}(\mathrm{x}_0)\right] &\leq \mathbb{E}_{q}\left[D_{\mathrm{KL}}(q(\mathrm{x}_T \mid \mathrm{x}_0)\|p(\mathrm{x}_T))\right] \\ &+ \sum_{t>1} D_{\mathrm{KL}}(q(\mathrm{x}_{t-1} \mid \mathrm{x}_t, \mathrm{x}_0)\|p_\theta(\mathrm{x}_{t-1} \mid \mathrm{x}_t)) - \log p_\theta(\mathrm{x}_0 \mid \mathrm{x}_1) \end{aligned} \end{equation}

where $L_T$ and $L_0$ represent the prior term and the reconstruction loss, respectively, and $\mathcal{L}_{1:T-1}$ represents the sum of KL divergence between the forward and backward steps. After simplifying $L_{t-1}$, the resulting posterior distribution can be expressed as:

\begin{equation} 
\label{deqn_ex1a9} 
q(\mathrm{x}_{t-1} \mid \mathrm{x}_t, \mathrm{x}_0) = \mathcal{N}\left(\mathrm{x}_{t-1}; \widetilde{\mu}_t(\mathrm{x}_t, \mathrm{x}_0), \widetilde{\beta}_t I\right) 
\end{equation}

Here, the term $\widetilde{\beta}_t$ is determined as a function dependent on $\beta_t$.
 Under this parameterization, the loss $\mathcal{L}_{t-1}$ can be further expressed as an expected $\ell_2$ loss by rewriting $\mathrm{x}_t$ as a function of $\mathrm{x}_0$ and noise:

\begin{equation} \label{deqn_ex1a10} \mathcal{L}_{t-1} = \mathbb{E}_{q}\left[\frac{1}{2\sigma_t^2} \left\| \widetilde{\mu}_t(\mathrm{x}_t, \mathrm{x}_0) - \mu_\theta(\mathrm{x}_t, t) \right\|^2 \right] + C \end{equation}

This form is related to denoising score matching (which will be discussed further in the next section). When $\mu_\theta$ is reparameterized as $\epsilon_\theta$, the training objective can be simplified to:

\begin{equation} \label{deqn_ex1a11} \mathcal{L}_{\mathrm{simple}} := \mathbb{E}_{\mathrm{x}_0, \boldsymbol{\epsilon}}\left[\frac{\beta_t^2}{2\sigma_T^2 \alpha_t (1 - \bar{\alpha}_t)} \left\| \epsilon - \epsilon_\theta\left(\sqrt{\bar{\alpha}_t} \,\mathrm{x}_0 + \sqrt{1 - \bar{\alpha}_t} \,\epsilon, t \right) \right\|^2 \right] \end{equation}

Currently, most diffusion models use this simplified DDPM training method. In subsequent improvements, $\mathcal{L}_{\mathrm{simple}}$ is often combined with other objective functions. After training, the noise prediction network $\epsilon_\theta$ will be applied to the reverse process to generate new data samples.

\subsection{Extensions and Improvements of Diffusion Models}

Since the proposal of DDPM, diffusion models have been continuously developed and improved in terms of theoretical framework, network structure, and practical application. Researchers have proposed a variety of improvement methods to solve the problems of slow generation speed and insufficient conditional controllability. These improvements include more efficient sampling mechanisms (such as denoising diffusion implicit models, DDIM)~\cite{46}, more robust probabilistic modeling methods (such as score-based models)~\cite{47}, stronger controllable generation strategies (such as conditional diffusion models)~\cite{48}, and the introduction of more expressive architectures such as Transformer. Owing to these developments, diffusion models are now markedly easier to apply and can be integrated into agricultural workflows with significantly greater efficiency.

\subsubsection{Denoising Diffusion Implicit Models (DDIM)}  

While models like DDPM~\cite{35,44} are capable of producing high-quality outputs, the requirement of many iterative denoising steps results in low computational efficiency, which poses a major obstacle for real-time use in agricultural scenarios.  
To solve this problem, DDIM~\cite{36} redesigned the diffusion path, which greatly improved the sampling speed without changing the core training process.

\begin{figure}[htbp] 
\centering 
\includegraphics[width=0.7\textwidth]{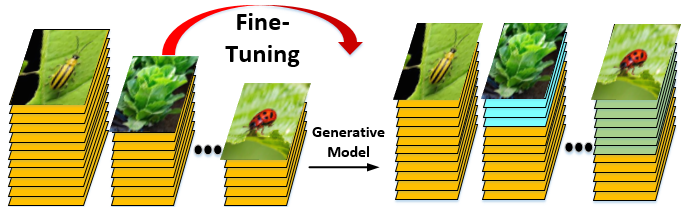} 
\caption{Diagram showing how dataset balancing is carried out. \cite{96}} 
\label{image4} 
\end{figure}

As shown in Figure~\ref{image4}, DDIM can accelerate image generation while maintaining the output quality in the long-tailed pest classification task, which is better than the traditional model. Unlike DDPM, which requires a large number of denoising steps to generate samples, DDIM skips some intermediate steps in the sampling trajectory, thus achieving faster and more efficient generation. In many pest detection tasks, the training data distribution is highly skewed, with some pest categories appearing only rarely and thus being significantly underrepresented.  
 Introducing artificially generated samples helps to alleviate this limitation by expanding minority categories and equalizing the dataset distribution. With a more balanced training set, models are able to learn more effectively, resulting in improved accuracy and stronger resilience in real-world applications.

DDIM proposes a more efficient sampling method on the basis of DDPM, which significantly accelerates image generation and does not need to modify the original training process. Unlike the random Markov chain method of DDPM, DDIM constructs a non-Markov process, and realizes deterministic sampling while keeping the marginal distribution of the original model unchanged.

This new form allows some steps to be skipped in the reverse generation process, increasing the sampling speed by 10 to 50 times without losing image quality. Because the edge distribution remains consistent, the trained DDPM model can be directly reused without retraining, which greatly enhances compatibility. These features make DDIM very attractive in large-scale applications, especially in agricultural scenarios where efficiency is high.

The improvement of DDIM is that it introduces the idea of "implicit" modeling in the denoising process. Traditional diffusion models need to generate the next image step by step through the Markov chain. Although the quality is high, the calculation cost is expensive, especially when a large number of denoising steps are required. The core innovation of DDIM is the implicit modeling of the reverse process. It no longer requires explicit noise removal in each step, but allows the image to be generated in fewer steps through the improved reverse reasoning process. DDIM introduces a parameter to modulate denoising intensity, giving the reverse process more flexibility and breaking away from linear transitions, which enhances image quality.  

Its reverse step is given by:  

\begin{equation} 
p_\theta(x_{t-1} \mid x_t) = \mathcal{N}(x_{t-1}; \mu_\theta(x_t, t), \sigma_t^2 I) 
\end{equation}
In this formulation, $\mu_\theta(\mathrm{x}_t, t)$ denotes the predicted mean, and $\sigma_t$ specifies the noise level.

This formula achieves an efficient reverse generation process by reasonably selecting the noise injection method and generation parameters.

Unlike traditional diffusion models, DDIM does not need to traverse the entire Markov chain, and its implicit modeling method enables high-quality samples to be generated in fewer steps. This is due to the simplified generation steps and flexible denoising methods. By introducing the implicit reverse process, DDIM effectively reduces the computational overhead of the traditional diffusion framework, and realizes the high-quality image synthesis in fewer steps, which is very suitable for real-time and large-scale agricultural applications.


\subsubsection{Score-based Models (SDE)}

Score-based models~\cite{47} have recently risen to prominence as a distinctive type of generative framework. Instead of directly approximating the data distribution, they exploit the score function—defined as the gradient of the log-density—to steer the sampling process. This reformulation departs radically from classical paradigms like GANs and VAEs, and provides a mathematically principled route to data generation. By building the generative process on a gradient-driven dynamic mechanism, these models have proven to be efficient tools and have achieved breakthrough progress in tasks such as realistic image synthesis, super-resolution, and content restoration.
Rather than following the paths of earlier generative approaches such as generative adversarial networks (GANs)~\cite{28,32} or variational autoencoders (VAEs)~\cite{49}, these models establish a distinct paradigm built on rigorous theoretical underpinnings. By leveraging this framework, score-based models have delivered outstanding results across a range of applications, including image generation~\cite{50}, super-resolution~\cite{51,52}, and image inpainting~\cite{53,54}.

The goal of generative models is to generate new samples from the target data distribution. Early models such as VAE and GAN attempt to directly model the data distribution, or achieve it through adversarial training, but often face problems such as instability and mode collapse. In contrast, score-based models generate samples by estimating the score function. The score function corresponds to the gradient of the log probability density and reveals how a sample should be adjusted so that it better conforms to the underlying data distribution. If $x \sim p_{\text{data}}(x)$, then the score function is defined as

\begin{equation} s_\theta(x) = \nabla_x \log p_{\text{data}}(x) \end{equation}

It depicts the trend of probability density with the change of input, and provides directional guidance for sample generation.
However, directly computing $\nabla_x \log p_{\text{data}}(x)$ is often intractable because the true data distribution is unknown.

For this purpose, score matching was introduced as a training strategy. The idea is to minimize the difference between the score function estimated by the model and the real score function, and use the data samples after adding noise to approximate the gradient. Its loss function is generally expressed as:

\begin{equation} \mathcal{L}_{\text{SM}} = \mathbb{E}_{x \sim p_{\text{data}}} \left[ \left\| s_\theta(x) - \nabla_x \log p_{\text{data}}(x) \right\|^2 \right] \end{equation}

By training a model to approximate these gradients, we obtain a function that can be used to iteratively update and generate samples along the high-density direction. Minimizing this loss allows the model to learn an accurate approximation of the score function.

In practice, the score function is usually parameterized by a neural network, and the model is trained by adding noise to learn the denoising behavior under different perturbations, so as to cover multiple noise levels.

The generation process generally adopts the gradient ascent method. Starting from a Gaussian noise vector $x_T \sim \mathcal{N}(0, I)$, the sample is iteratively optimized into structured data:

\begin{equation} \label{deqn_ex1a15} x_{t-1} = x_t + \Delta t \cdot s_\theta(x_t) \end{equation}

where $\Delta t$ is the step size and $s_\theta(x_t)$ provides the update direction. through continuous iteration, the sample is gradually transformed from pure noise to structured data.

\begin{figure}[htbp] 
\centering 
\includegraphics[width=0.7\textwidth]{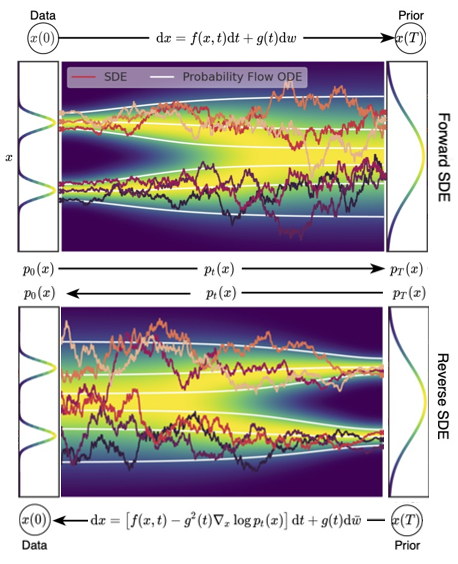} 
\caption{SDE-driven score-based generative model \cite{47}} 
\label{image5} 
\end{figure}

The key to the success of score models is the use of multiple noise scales, and this idea is further extended to continuous infinite noise levels, so that the perturbed data distribution can be evolved through stochastic differential equations (SDE). The general architecture is illustrated in Figure~\ref{image5}.

From a broader perspective, score-based models have redefined the paradigm of generative learning. Instead of directly reconstructing the complete data distribution, they use the gradient of the log density to guide the optimization of samples. This shift has brought many advantages: higher quality of generated results, more stable training process, and a more solid theoretical foundation. In addition, this method is naturally compatible with diffusion models, providing new ideas and directions for the development of generative modeling.

By combining score functions with diffusion models, score-based models have shown broad application prospects in tasks such as image generation and restoration. As research continues to deepen, score-based models are expected to play an increasingly important role in more generative tasks.

\subsubsection{Noise Conditional Score Network (NCSN)}

Score-based Generative Model (SGM), also known as Noise-Conditioned Score Network (NCSN)~\cite{37,55,56}, was proposed by Yang Song et al. in 2019. The main goal is to generate high-quality data by learning the score function at multiple noise levels.  
The score function is the gradient of the log probability density.  

For a data distribution $p(x)$, it is defined as:  

\begin{equation} 
\label{deqn_ex1a13} 
s(x) = \nabla_x \log p(x) 
\end{equation}  

It reveals how a data point should be shifted within the input space. 
It guides the point toward regions of higher probability.  
To estimate this function, a neural network $s_\theta(x)$ controlled by parameters $\theta$ is typically trained to approximate the true score function, with the following optimization objective:

\begin{equation} \label{deqn_ex1a16} \mathbb{E}_{x \sim p(x)} \left\| s_\theta(x) - \nabla_x \log p(x) \right\|_2^2 \end{equation}

However, it is difficult to directly calculate $\nabla_x \log p(x)$ in high-dimensional data, so the traditional score matching method is not applicable. For this reason, Song et al.~\cite{37} introduced denoising score matching~\cite{57} and sliced score matching~\cite{58}.

Another challenge is the error in score estimation in low-density regions. Since real data is often located on low-dimensional manifolds in high-dimensional space (manifold hypothesis), the authors propose adding multi-scale Gaussian noise during training to smooth the data distribution and improve the score learning effect.

To handle different noise levels, a single noise-conditioned score network is trained. Specifically, for Gaussian noise distribution $p_{\sigma_t}(x_t \mid x) = \mathcal{N}(x_t; x, \sigma_t^2 I)$, the score function is:

\[ \nabla_{x_t} \log p_{\sigma_t}(x_t \mid x) = -\frac{x_t - x}{\sigma_t} \]

Therefore, for a set of noise levels $\sigma_1 < \sigma_2 < \cdots < \sigma_T$, Eq.~\ref{deqn_ex1a13} can be transformed into the following training objective:

\begin{equation} \label{deqn_ex1a17} \frac{1}{T} \sum_{t=1}^T \lambda(\sigma_t) \, \mathbb{E}_{x \sim p(x)} \, \mathbb{E}_{x_t \sim p_{\sigma_t}(x_t \mid x)} \left\| s_\theta(x_t, \sigma_t) + \frac{x_t - x}{\sigma_t} \right\|_2^2 \end{equation}

where $\lambda(\sigma_t)$ is a weighting function to balance the contribution of different noise scales.

When sampling from the target distribution $p(x)$, the Langevin dynamics iterative method is used. The initial sample $x_0 \sim \pi(x)$ and the update rule is:

\begin{equation} \label{deqn_ex1a18} x_i = x_{i-1} + \frac{\gamma}{2} \nabla_x \log p(x) + \sqrt{\gamma} \cdot \omega_i \end{equation}

where $\omega_i \sim \mathcal{N}(0, I)$, $i \in \{1, \ldots, N\}$. When $\gamma \to 0$ and $N \to \infty$, the sequence $\{x_i\}$ converges to the target distribution $p(x)$.

To improve robustness and avoid common failure modes, Song et al.~\cite{35} proposed the \textit{annealed Langevin dynamics} algorithm. This method gradually reduces the noise scale $\sigma_i$ (i.e., annealing) during the sampling process, thereby improving the sample quality and training stability~\cite{55}.

\subsubsection{Conditional Diffusion Models (CDMs)}

The basic idea of the diffusion model is to learn to recover clean samples from random noise by gradually adding noise to the data and training the model to remove the noise in reverse. This training method is different from adversarial methods such as GAN, and is more stable, and the generation quality is more consistent.

Based on this framework, conditional diffusion models (CDMs) introduce additional contextual information (such as category labels, text descriptions, or reference images) to guide the generation process. With the help of conditional information, CDMs can achieve controllable generation, and are often used in tasks such as text-to-image generation, speech generation, and image restoration.

In mathematical form, these models can sample from either an unconditional distribution $p_0$ or a conditional distribution $p_0(\mathrm{x}|c)$, where $c$ represents the conditional information associated with the input $x$ \cite{59}. This flexibility has driven the rapid development of controllable generation. To achieve conditional generation, several sampling methods have been proposed, including classifier-free guidance \cite{60} and classifier-based guidance sampling \cite{61}.

\textbf{Label-based condition:} The generation is guided by introducing conditional gradients during the sampling process. Generally, an additional classifier is required, and the U-Net coding structure is commonly used to generate conditional gradients. Conditional forms include text, categories, binary indicators, or data features \cite{61,62,63,64,65,66,67,68,69,70}. This method was first proposed by \cite{61} and has now become one of the basic technologies for conditional sampling.

\textbf{Unlabeled condition:} Guidance is provided using intrinsic or self-generated information, often applied in a self-supervised manner \cite{71,72}. This approach is often used for denoising \cite{73}, image-to-painting translation \cite{74}, and image inpainting \cite{75}.

\subsection{Relationship to Other Generative Models}

Diffusion models are related to and different from other generative models. First, there are likelihood-based methods, and second, there are GANs.

Diffusion models share similarities with variational autoencoders (VAEs): both map data to a latent space and reconstruct data from latent representations; both have objective functions that can be viewed as lower bounds on the data likelihood. But there are also significant differences. The latent representation of VAE is a compressed form of the original image, while the diffusion model gradually transforms the data into pure noise through forward diffusion. The latent space of VAE has lower dimensions, which is more conducive to efficient training; the diffusion model maintains the same dimensions as the original data, and the latent space is not trainable, but is obtained only through gradual noise addition. Recent studies have attempted to apply diffusion models to the latent space of VAEs to improve performance \cite{76,77}.

Autoregressive models \cite{78,79} treat data as a sequence and generate it step by step, with each pixel depending on the previous pixels. Although effective, unidirectional generation has limitations. Esser et al. \cite{80} proposed a hybrid model combining diffusion and autoregression, using an autoregressive network to achieve each step of transition in the diffusion process, so as to make up for the deficiency of one-way generation by using the global context.

Normalizing Flows \cite{81,82} map Gaussian distributions to complex data distributions through a series of reversible and differentiable transformations. Because its Jacobian determinant is easy to calculate, the likelihood can be accurately calculated. Similar to diffusion models, flow models also map between Gaussian and data distributions, but they are deterministic and reversible, while diffusion is random and gradual. The DiffFlow method \cite{83} combines the two, with both reversibility and generation quality.

Energy-based models (EBMs) \cite{84,85,86,87} represent data distributions with energy functions, which are flexible but difficult to train and require complex sampling such as MCMC. The training and sampling of diffusion models can be regarded as a special case of EBM, especially when based on score functions.

Until recently, GANs \cite{88} have been the dominant approach to generative modeling, especially in terms of image quality. However, its adversarial training brings instability and mode collapse \cite{89}. In contrast, diffusion models are more stable to train and have stronger sample diversity, but their inference efficiency is low and requires multiple network calculations.

In the latent space, GANs usually operate in a low-dimensional latent space and contain semantic directions that can be used for attribute manipulation \cite{28,90}. Diffusion models, on the other hand, maintain the same dimension as the input and rely more on external guidance (such as text prompts) to achieve controllability. Song et al. \cite{47} showed that interpolation in the latent space of diffusion models can generate smooth transition images, showing a potential research direction.

\begin{figure}[htbp] \centering \includegraphics[width=0.7\textwidth]{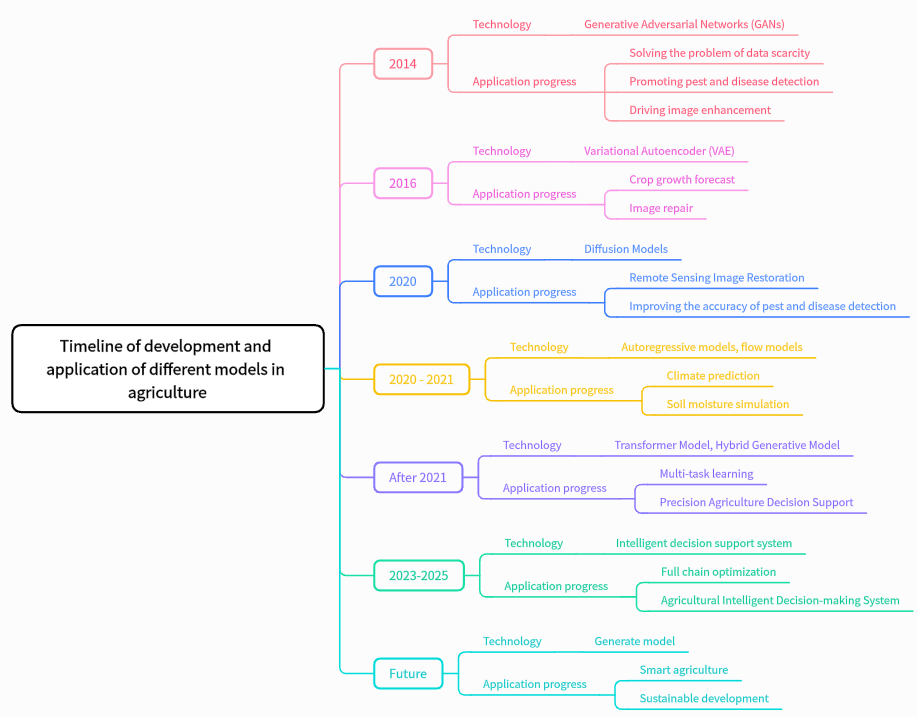} \hfil \caption{The evolution of different models in agriculture.} \label{image6} \end{figure}

In summary, diffusion models have both commonalities and differences with GANs, VAEs, flow models, and EBMs in terms of generation mechanism, training method, and latent space structure. They have advantages in training stability and sample diversity, can avoid mode collapse, and their performance is often better than GAN. However, the inference cost is still high. Future research can focus on combining the advantages of different generation paradigms, improving inference efficiency, and deeply exploring the semantic structure of the latent space of diffusion models. As shown in Figure~\ref{image6}, the application evolution of different models in the agricultural field is summarized; Table~\ref{tab1} compares their performance in agricultural tasks.

\begin{table}[htbp] \centering \caption{Performance comparison of different generative models in agricultural tasks} \label{tab1} \begin{tabularx}{\textwidth}{C C C C C} \toprule \textbf{Feature} & \textbf{Diffusion model} \cite{35,44} & \textbf{Adversarial network (GAN)} \cite{28,32} & \textbf{Autoencoder (VAE)} \cite{49} & \textbf{Normalization flow} \cite{81,82} \\ \midrule

\bfseries Generation mechanism & Gradual denoising to generate data & Adversarial training of generator and discriminator & Encoder-decoder generates data & Map distribution through reversible transformation \\ \midrule

\bfseries Training stability & Stable, avoids mode collapse & Prone to mode collapse & Stable, but lower generation quality & High stability, but high computational cost \\ \midrule

\bfseries Generation quality & High quality, strong diversity & Unstable quality, may lack diversity & General quality, poor performance under complex distribution & Low quality, suitable for simple distribution \\ \midrule

\bfseries Computational efficiency & Low, multiple network calculations are required & High, but easy to fall into local optimum & High efficiency, suitable for large-scale data & High computational cost, but strong expression ability \\ \midrule

\bfseries Pattern collapse problem & Pattern collapse can be avoided & Pattern collapse is prone to occur & No obvious problem & No obvious problem \\ \midrule

\bfseries Applicability & Suitable for high-quality generation tasks & Suitable for tasks such as images and videos & Suitable for generation, representation learning and dimensionality reduction & Suitable for high-dimensional data modeling \\ \bottomrule \end{tabularx} \end{table}


\section{Application of Diffusion Model in Agricultural Image Processing and Analysis}\label{sec4}

This section introduces the latest advances of diffusion models in agricultural image processing and divides its applications into seven aspects, as shown in Figure~\ref{image12}. In recent years, generative models, especially diffusion models, have shown great potential in the agricultural field, and typical applications include crop health monitoring \cite{107,108,91}, pest and disease identification \cite{113,18,109,110,112}, and decision support systems \cite{16}. By achieving high-quality image generation, denoising, enhancement, and data augmentation, diffusion models have effectively improved the accuracy of crop monitoring and disease detection, generated diverse agricultural images \cite{39,91,94,96}, expanded training datasets \cite{15,92}, and enhanced the performance of deep learning models.

\subsection{Image Classification}

Diffusion models are becoming increasingly important in agricultural image classification, especially in dealing with insufficient data, class imbalance, and improving model generalization, such as plant disease detection \cite{107,92,93} and crop growth monitoring \cite{98}.Agricultural image datasets usually lack sufficient samples and show class imbalance.  
By producing diverse and realistic synthetic images, diffusion models expand the training data.  
This expansion strengthens classifiers and boosts their performance.  
 The realistic data generated by it not only improves the classification accuracy, but also enhances the adaptability in different environments, and is an important tool for smart agriculture.

\begin{figure}[htbp] \centering \includegraphics[width=0.7\textwidth]{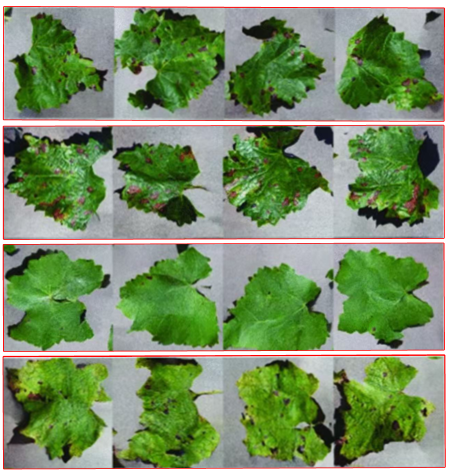} \caption{Images generated by EDM.\cite{93}} \label{image7} \end{figure}

In order to improve the accuracy of sunflower disease classification, Zhou et al. \cite{107} proposed a few-shot learning framework based on diffusion model, combined diffusion generation with few-shot learning, and constructed an efficient end-to-end classification system. A novel image generation method based on DDPM is proposed by  Wang et al \cite{92} who introduced the ECA attention mechanism to address the problem of insufficient data for ginseng leaf disease identification. They combined Inception-SSNet, skip connections, feature fusion, and ScConv convolution to improve the recognition effect of small lesions and similar diseases. Zhang et al. \cite{93} combined the diffusion model (EDM) with data augmentation in the classification of grape leaf diseases, used EDM to generate synthetic images to alleviate the category imbalance, and used the augmented data to train CNN (ResNet50, VGG16, AlexNet), which significantly improved the classification performance. Figure~\ref{image7} shows the images generated by EDM.

Du et al. \cite{96} proposed the ADM-DDIM framework to generate high-quality pest images, and experiments showed that synthetic data significantly improved the performance of the classification network. Lee~\cite{98} applied DDPM for latent space interpolation.  
By interpolating between normal and wilted images, new samples were created.  
These additional samples improved the accuracy of classification.  

\subsection{Image Generation}

Diffusion models also show strong capability in image generation.  
They are widely used for crop image synthesis and restoration \cite{39,94}.  
Typical applications include missing region completion and growth stage simulation \cite{91,143}.  
 These high-quality images not only improve visual analysis, but also assist agronomists in state assessment, pest and disease prediction, and growth environment simulation, providing technical support for precision agriculture.

\begin{figure}[htbp] \centering \includegraphics[width=0.7\textwidth]{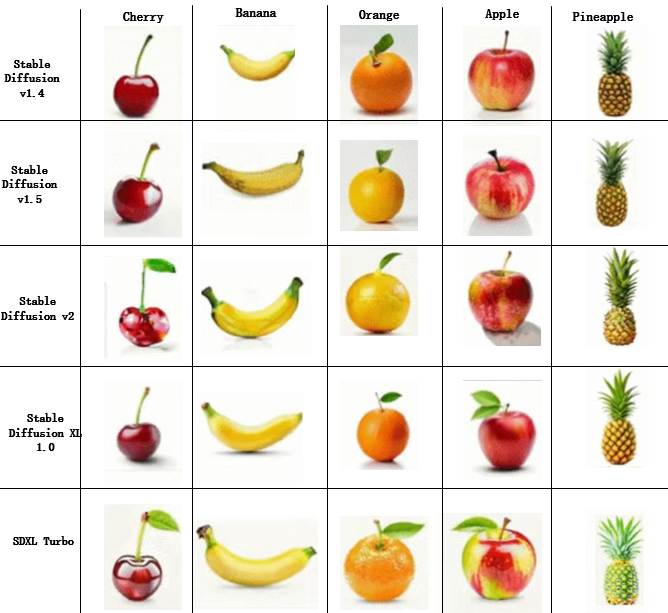} \caption{Generation examples of different fruits. \cite{39}} \label{image8} \end{figure}

In precision agriculture, Zhao et al. \cite{39} used the Stable Diffusion model (v1.4, v1.5, v2, SDXL) to construct a synthetic fruit image dataset, as shown in Figure~\ref{image8}.Synthetic images of various fruits—such as cherries, bananas, oranges, apples, and pineapples—were produced.  
This greatly enriched the dataset.  
It also led to a clear improvement in model accuracy.  
SDXL Turbo performed best in all categories, showing the potential of generative AI in fruit detection and segmentation for agricultural robots.

Plant diseases are a serious threat to agricultural production. Traditional methods rely on manual labor and expert experience, which are time-consuming, labor-intensive, and prone to errors. Generative models (especially Stable Diffusion) provide an effective way to alleviate data insufficiency and class imbalance. Huang et al.~\cite{91} applied Stable Diffusion to create high-quality disease images.  
They further used ControlNet and LoRA fine-tuning.  
The dataset became more diverse, and the model achieved better generalization.  
Heschl et al.~\cite{94} introduced a dual diffusion framework that integrates DDPM with GANs.  
Their approach produced realistic phenotypic traits together with consistent labels.  
As a result, both the precision and efficiency of crop area segmentation were improved.

To improve weed recognition and expand the dataset, Chen et al. \cite{143}, Deng et al. \cite{144} and Moreno et al. \cite{145} explored data augmentation based on diffusion models. Chen et al. used Stable Diffusion to generate weed images and combined them with CNN, which significantly improved the detection effect in complex scenes. Deng et al. reported that mixing real images with synthetic ones enhanced classification,  
especially for overlapping weed species.  
Moreno et al. verified the use of Stable Diffusion for data synthesis.  
With this approach, detection became more accurate and more robust.  
The findings provide strong evidence for the value of diffusion-based augmentation.

Du et al.~\cite{96} focused on rare classes in pest datasets and introduced diffusion models.  
Their method adjusted the data distribution and enhanced detection accuracy.  
Realistic pest samples were created by fine-tuning a pre-trained model.  
These synthetic images expanded the dataset, increasing both diversity and quality.  
On the IP102 benchmark, the improvements were evident.  
Diffusion models surpassed conventional methods in generating balanced and high-quality data.  
They effectively mitigated the long-tail challenge and further raised classification performance.  

\subsection{Data Augmentation}

Generative modeling has progressed quickly in recent years.  
Among its many approaches, diffusion models have become a central force driving this advancement.  
Agriculture, in particular, has benefited greatly from these developments.  
In data augmentation, they contribute to crop monitoring \cite{98,142} and disease detection \cite{92,93} through the production of diverse, high-quality synthetic data.  
By preserving semantic meaning while introducing variability, they expand dataset diversity.  
This allows models to better recognize crop conditions as well as pest and disease categories.  
As a result, under limited-resource settings, diffusion models provide valuable data support and greatly enhance model generalization.

\begin{figure}[htbp] \centering \includegraphics[width=0.7\textwidth]{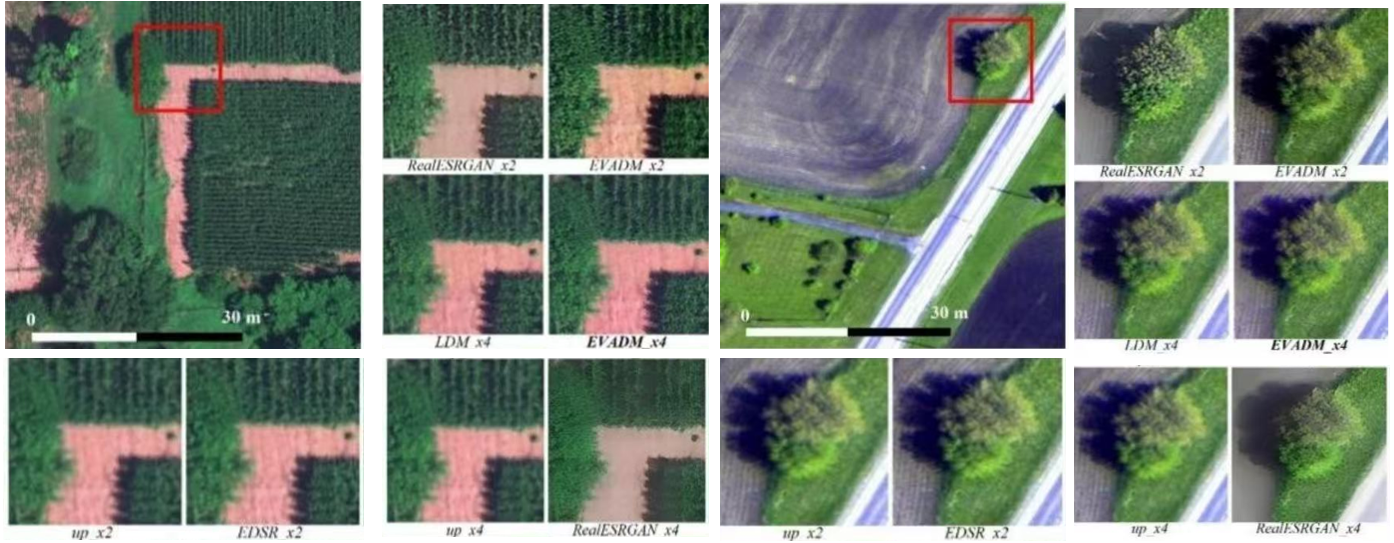} \caption{Comparison of Super-Resolution Results on the Agricultural Vision (AgV) Datasets. The left column shows the Field of View (FOV) of the AgV data sample, and the right column is the SR result. EVADM generates more realistic and detailed views. \cite{95}} \label{image9} \end{figure}

Image super-resolution (SR) is a key technology to improve the resolution and quality of remote sensing images. In agricultural remote sensing analysis, SR helps to enhance details and interpretability. Stable Diffusion performs well in agricultural super-resolution tasks. Lu et al. \cite{95} constructed the CropSR dataset and applied the VASA and EVADM models, as shown in Figure~\ref{image9}. The results show that the diffusion model significantly improves the image quality in the SR task, and has advantages in indicators such as Fréchet Inception Distance (FID) and Super-Resolution Fidelity Index (SRFI), showing its potential in agricultural image enhancement.

In plant disease classification, insufficient data and category imbalance are common bottlenecks. Taking ginseng leaf disease as an example, the limited number of samples and variability affect the classification accuracy. Wang et al.~\cite{92} introduced a data augmentation approach built on diffusion models.  
Their method produced realistic synthetic images of high quality.  
The generated samples expanded the diversity of the dataset.  
As a result, the overall performance of the model was enhanced.   
Zhang et al. \cite{93} proposed the Elucidated Diffusion Model (EDM), which generates diverse disease images by sampling at different noise levels, and is used to train models such as ResNet50, VGG16, and AlexNet, which significantly improves the classification accuracy and is especially suitable for identifying subtle diseases. Hirahara et al. \cite{97} proposed a diffusion-based enhancement method D4, which uses a text-guided diffusion model to generate labeled vineyard branch detection images, which not only match the target domain but also retain the labeled information, thereby improving the detection performance and alleviating the labeling difficulty and domain adaptation problems.

\begin{figure}[htbp] \centering \includegraphics[width=0.7\textwidth]{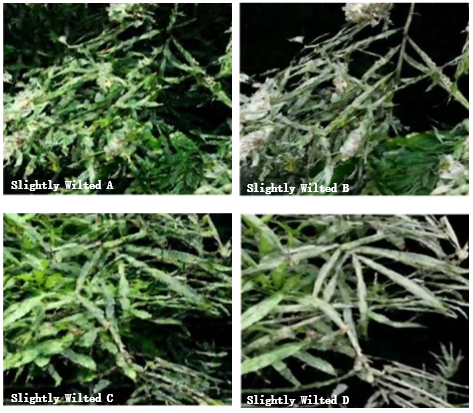} \caption{The results of generating synthetic data of "mild wilting" using the diffusion model. \cite{98}} \label{image10} \end{figure}

In the early detection of diseases, traditional image enhancement methods are often difficult to simulate subtle differences. Lee \cite{98} proposed a DDPM-based method to generate images of intermediate states such as "mild wilting". As shown in Figure~\ref{image10}, these synthetic images improve the classification accuracy, enhance the sensitivity to the health status of plants, and improve the robustness of the monitoring system.

To enable real-time recognition of broken grains and impurities in harvesting, Zhang et al.~\cite{142} designed the Dual Attention Diffusion Model (DADM).  
This framework fuses the denoising diffusion probabilistic model (DDPM) with a spatial channel attention block (SCA-Block).  
By doing so, it generates images with improved fidelity and richer diversity compared to both DDPM and GAN.  
The resulting synthetic data enhanced the accuracy of seven segmentation networks and further demonstrated strong transferability to tasks such as pest and disease identification.  


\subsection{Remote Sensing and Hyperspectral Image Reconstruction}

With the development of remote sensing technology, the reconstruction and enhancement of remote sensing images, especially high-resolution and hyperspectral data, have become key tasks in environmental monitoring, agricultural planning and disaster response. Traditional reconstruction methods, such as interpolation \cite{99,100} or CNN-based enhancement \cite{101}, are often limited in the case of large-scale data loss or severe degradation. In contrast, diffusion models have shown great potential in the reconstruction of complex agricultural remote sensing images.

Satellite images often have missing data due to cloud cover, sensor failure, or incomplete acquisition, especially in high-resolution and high-frequency observations. Diffusion models can recover these data through progressive denoising. Yu et al. proposed the SatelliteMaker \cite{102} framework to reconstruct spatial and spatiotemporal gaps in remote sensing data using diffusion models. This method combines the digital elevation model (DEM) with custom prompts, and performs well in agricultural monitoring. The diffusion model with latent structure combined with AdamW optimizer and Huber loss function \cite{103} performs well in cloud removal tasks and can maintain surface details.

Hyperspectral images (HSIs) are widely used in crop health assessment and soil analysis due to their rich spectral information. However, due to hardware limitations, the resolution is often low. Diffusion models can convert low-resolution HSIs into high-resolution images through progressive denoising and refinement. Tang et al. \cite{41} applied the EVADM model to the CropSR dataset, which significantly improved the resolution and details, and provided a new method for crop monitoring and spectral analysis.

Land use and land cover (LULC) semantic segmentation is also a key application. Although CNN and visual Transformer are widely used, they are still insufficient in capturing details of high-resolution images. Fan et al. proposed DDPM-SegFormer \cite{104}, which combines DDPM with ViT to achieve more refined multi-scale semantic extraction and detail retention, and is superior to SegFormer and SegNext on the LoveDA and Tarim Basin datasets.

Diffusion models also perform well in agricultural aerial image super-resolution tasks. Lu et al. \cite{95} proposed a method combining diffusion model, variance attention mechanism and self-supervised training, which significantly improved the quality of aerial images. Han et al. proposed EHC-DMSR \cite{105}, which integrates CNN and Transformer features and introduces frequency constraints to improve the super-resolution effect while reducing the inference time. Graikos et al. \cite{106} proposed a self-supervised diffusion model for generating diverse remote sensing images and improving the classification and LULC mapping effects.

\subsection{Diagnosis and Auxiliary Decision-making of Diseases and Pests}

Diffusion models also show advantages in pest and disease detection, especially in feature extraction, lesion localization, background suppression and decision support.

Yin et al. \cite{16} proposed a diffusion-based plant disease detection method, which introduced an endogenous diffusion sub-network to optimize the feature distribution and improve the lesion recognition ability in complex backgrounds, especially suitable for fine-grained situations such as rust and root rot. The proposed endogenous diffusion loss can adaptively adjust the diffusion steps, realize multi-task learning and enhance robustness.

Huang et al. \cite{91} proposed an enhancement method based on Stable Diffusion, which combines ControlNet and low-rank adaptation (LoRA) to achieve precise control of the generation process, improve the FID score and Top-1 classification accuracy, and the effect is particularly prominent under small sample conditions.

\begin{figure}[htbp] \centering \includegraphics[width=0.7\textwidth]{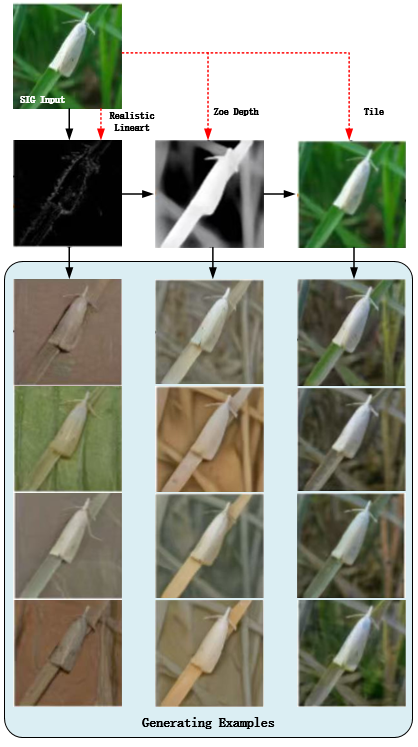} \caption{Example of image generation using the SIG Block. Semantic embeddings (Realistic Lineart, Zoe Depth, and Tile) are gradually integrated during the generation process.\cite{18}} \label{image11} \end{figure}

\begin{figure}[htbp] \centering \includegraphics[width=0.7\textheight]{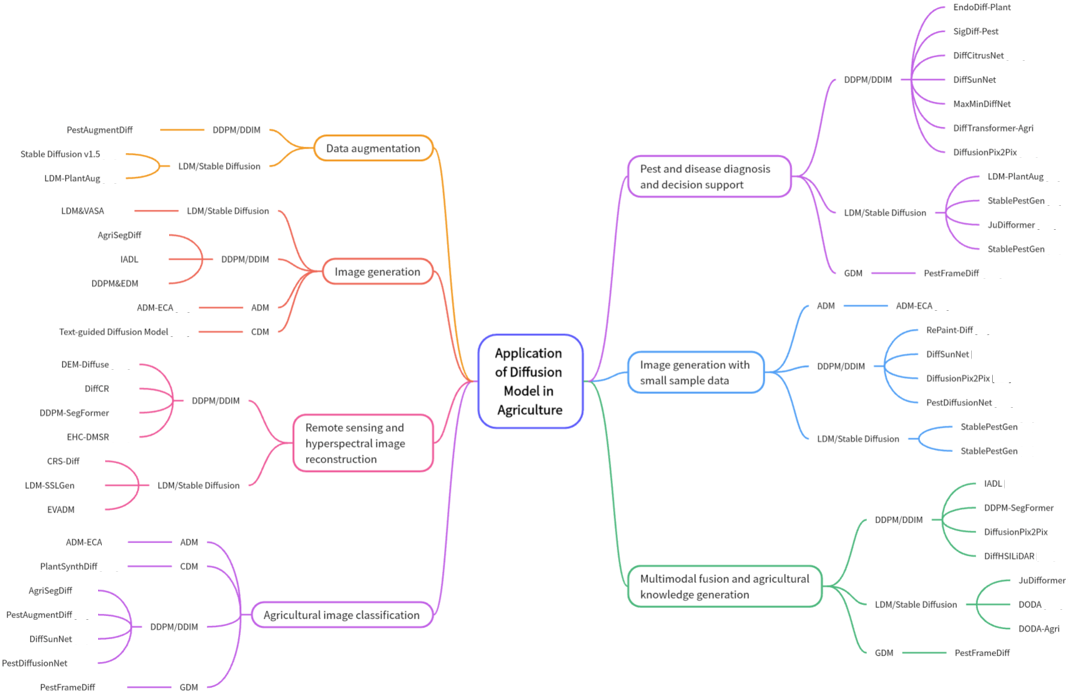} \caption{Seven application areas of diffusion models in agriculture: (1) data augmentation, (2) image generation, (3) remote sensing and hyperspectral image reconstruction, (4) pest and disease diagnosis and decision support, (5) small sample image generation, (6) multi-modal fusion and agricultural knowledge generation, and (7) agricultural image classification.} \label{image12} \end{figure}

Wang et al. \cite{18} proposed a method combining semantic diffusion and knowledge distillation to generate diversified pest images. Their SIG-MLAD framework improves the image quality and diversity through semantic guidance and multi-level alignment, and improves the detection effect under complex field conditions. As shown in Figure~\ref{image11}, their model is robust in real-time pest detection.

Li et al. \cite{109} applied DDPM and transfer learning to citrus disease diagnosis. They proposed two methods: one is to pre-train and fine-tune the Swin Transformer using synthetic data, with an accuracy of 96.3\%; the other is to combine synthetic data with the pre-trained Swin Transformer, with an accuracy of up to 99.8\%, which effectively solves the problem of small samples.

Guo et al. \cite{110} proposed a lightweight detection framework combining MaxMin diffusion mechanism and time series modeling, which improved the segmentation accuracy in complex scenes and performed well in tomato bacterial spot detection.

Hu et al. \cite{111} proposed a jujube disease detection framework based on Diffusion-Transformer. through the parallel attention mechanism and loss function, the accuracy of the model reached 95\%, which is better than the traditional method. Zhang et al. \cite{112} proposed a multi-modal deep model that combines image and sensor data and is implemented through diffusion and Transformer modules. Detecting jujube tree diseases in desert environments, the model is superior to the baseline method in terms of accuracy, recall rate and robustness, especially in low light and dusty conditions.


\subsection{Multi-modal Fusion and Agricultural Knowledge Generation}

In agricultural research, the fusion of multi-modal data is both important and challenging.
Diffusion models effectively address this need by integrating multi-source information \cite{20,104,112,27}, such as remote sensing images, sensor outputs, and environmental conditions.
through this fusion, the model can generate high-quality synthetic samples, thereby enriching the data set, improving the reliability of the model, and providing support for the generation of agricultural knowledge to serve decision-making and precise management.

Xiang et al. proposed a diffusion-based domain adaptation framework DODA \cite{115} to solve the adaptation problem of agricultural target detection models in dynamic environments. This method combines external domain embedding and synthetic images, which can quickly adapt to the new domain and promote multi-modal fusion. Kong et al. \cite{116} developed a disease detection framework that combines diffusion models, graph attention networks (ViG), and lightweight optimization techniques. Through multi-modal fusion, this method improves detection accuracy and robustness in complex environments, and performs exceptionally well in mobile applications (such as tomato disease diagnosis).

Fan et al.~\cite{104} introduced DDPM-SegFormer, a framework that unites the denoising diffusion probabilistic model (DDPM) with a vision transformer (ViT).  
This approach enables highly accurate land use and land cover (LULC) segmentation and demonstrates strong potential across a variety of agricultural remote sensing applications.  
Zhang et al. \cite{112} proposed the JuDifformer model, which uses the diffusion mechanism to fuse image and sensor data, improved disease detection effectiveness of jujube trees in desert environments, promoted the generation of agricultural knowledge, and optimized disease prediction and control.

Chen et al. \cite{27} proposed the IADL framework, which combines the diffusion model with natural language processing, guides image generation through text description, enhances the plant disease detection data set, improves the detection accuracy, and promotes the automatic generation of agricultural knowledge. Jiang et al.~\cite{117} introduced a framework based on DDPM that brings together hyperspectral imagery and LiDAR information.  
The integration of these two modalities resulted in significant improvements in both disease detection and crop monitoring performance.  

Its effectiveness was especially evident when applied to complex and demanding agricultural environments.  
 Zhang et al. \cite{118} developed a BO-CNN-BiLSTM model that integrates remote sensing and climate data to improve the accuracy of crop yield prediction and provide support for smart agriculture and sustainable development. Zhou et al. \cite{119} also proposed a multi-modal fusion and agricultural knowledge generation system based on JuDifformer, which provided a new path for precision agriculture in desert areas.

\subsection{Image Generation with Few-Shot Data}  

In agriculture, the scarcity of labeled data often limits the performance of detection models.  
Diffusion models address this challenge by producing realistic, high-quality images even from very small datasets.  
This ability is particularly important for strengthening pest and disease recognition.  
To overcome the shortage of labeled samples, recent research has increasingly coupled diffusion models with few-shot learning techniques.  
Such combinations allow models to maintain strong performance even when training data is scarce.

Zhou et al. \cite{107} proposed a few-shot learning framework based on diffusion model for soybean disease detection and classification. This framework improves the detection performance by generating high-quality synthetic samples, and improves the lesion feature extraction by using the attention mechanism. At the same time, a new diffusion loss function is proposed, and its effectiveness is verified in the laboratory and real scenes. In the case of soybean disease detection in Bayannur, China, the method achieved good results and showed practical value.

Wang et al. \cite{108} proposed a controllable generation framework based on Stable Diffusion, introduced text prompts and spatial coordinates, and synthesized realistic new pest images in the natural field environment. This method directly alleviates the problem of insufficient labeled data in the pest detection task and generates training samples that meet the scene.

Similarly, Astuti et al. \cite{113} constructed a synthetic potato leaf disease dataset based on Convolutional Vision Transformer (CvT) and Stable Diffusion. The CvT model was trained using synthetic images, and the detection accuracy was significantly improved. This shows the value of diffusion enhancement in agricultural data-limited scenarios.

In addition, Wang et al. \cite{92} and Egusquiza et al. \cite{114} also demonstrated the application of diffusion models in plant disease image enhancement and generation. For example, the RePaint model is superior to the traditional GAN in terms of image quality and disease category accuracy; DiffusionPix2Pix combines disease severity labels and leaf segmentation masks to generate high-quality disease images, which is superior to the traditional GAN in many indicators, and provides an efficient solution for disease data set enhancement and detection accuracy improvement.

\section{Experimental Results and Analysis}\label{sec5}

In order to further verify the performance advantages of diffusion models in agriculture, this paper designs and carries out multiple sets of comparative experiments. through data augmentation and image generation experiments, remote sensing and hyperspectral image reconstruction experiments, pest and disease diagnosis and decision support experiments, and multi-modal fusion experiments, the diffusion model and existing methods are comprehensively evaluated by visualization and quantitative indicators. This experiment focuses on the validity verification of diffusion model in agricultural pest image generation, and provides reference for subsequent research.

\subsection{Performance Evaluation Metrics}  

To evaluate the performance of generative and classification models, the metrics are summarized in Table~\ref{tab:metrics2}.

\begin{table*}[!ht]
\centering
\caption{Performance evaluation metrics used in this study}
\label{tab:metrics2}
\renewcommand{\arraystretch}{1.2}
\setlength{\tabcolsep}{6pt}
\begin{tabular}{p{6cm}p{9cm}}
\toprule
\textbf{Formula} & \textbf{Explanation } \\
\midrule
$\text{SSIM}(x,y)=\frac{(2\mu_x\mu_y+C_1)(2\sigma_{xy}+C_2)}{(\mu_x^2+\mu_y^2+C_1)(\sigma_x^2+\sigma_y^2+C_2)}$ & 
$\mu_x,\mu_y$: means; $\sigma_x,\sigma_y$: variances; $\sigma_{xy}$: covariance; $C_1,C_2$: stability constants.  
Structural similarity; closer to 1 means better detail fidelity. \\
\midrule
$\text{PSNR}=10\log_{10}\left(\frac{MAX_I^2}{MSE}\right)$ & 
$MAX_I$: maximum pixel value; $MSE$: mean squared error.  
Higher values indicate better reconstruction with fewer artifacts. \\
\midrule
$d^2=(\|\mu_r-\mu_g\|_2^2 + \mathrm{Tr}(\Sigma_r+\Sigma_g-2(\Sigma_r\Sigma_g)^{1/2}))$ & 
$\mu_r,\Sigma_r$: mean and covariance of real distribution; $\mu_g,\Sigma_g$: mean and covariance of generated distribution.  
Lower values mean generated images are closer to real. \\
\midrule
RFID & 
Relative FID against upscaled results. Same as FID, but reference distribution is from upscaled images.  
Used to compare generated vs. baseline upscaling. \\
\midrule
$\text{SRFI}=\alpha \cdot \text{SSIM} + \beta \cdot \text{Perceptual}$ & 
$\alpha,\beta$: weighting coefficients.  
Combines structural and perceptual measures; higher scores = clearer SR images. \\
\midrule
FLOPs  & Calculation the sum of convolution/matrix-multiplication operations. Total floating-point operations.  
Fewer FLOPs = more efficient model. \\
\midrule
Params &Total number of trainable parameters.
Weights and biases in the model.  
Fewer parameters = lighter and easier to deploy. \\
\midrule
$Precision=\frac{TP}{TP+FP}$ & 
$TP$: true positives; $FP$: false positives.  
Precision: higher = fewer false alarms. \\
\midrule
$Recall=\frac{TP}{TP+FN}$ & 
$FN$: false negatives.  
Recall: higher = fewer missed detections. \\
\midrule
$Accuracy=\frac{TP+TN}{TP+TN+FP+FN}$ & 
$TN$: true negatives.  
Accuracy: overall correctness. \\
\midrule
$\frac{1}{N}\sum_{i=1}^N AP_i(\text{IoU}=0.75)$ & 
$AP_i$: average precision for class $i$.  
mAP@75: higher = better detection/localization. \\
\midrule
$2 \cdot \frac{Precision \cdot Recall}{Precision+Recall}$ & 
F1-Score: balances precision and recall, especially under imbalance. \\
\midrule
$\frac{\text{frames}}{\text{second}}$ & 
FPS: frames per second.  
Higher = stronger real-time performance. \\
\midrule
GPU memory usage during inference & 
Lower usage = better for resource-limited devices. \\
\bottomrule
\end{tabular}
\end{table*}

\subsection{Experimental Results} \subsubsection{Data Augmentation and Image Generation Experiments}

\begin{figure}[htbp] \centering \includegraphics[width=0.7\textwidth]{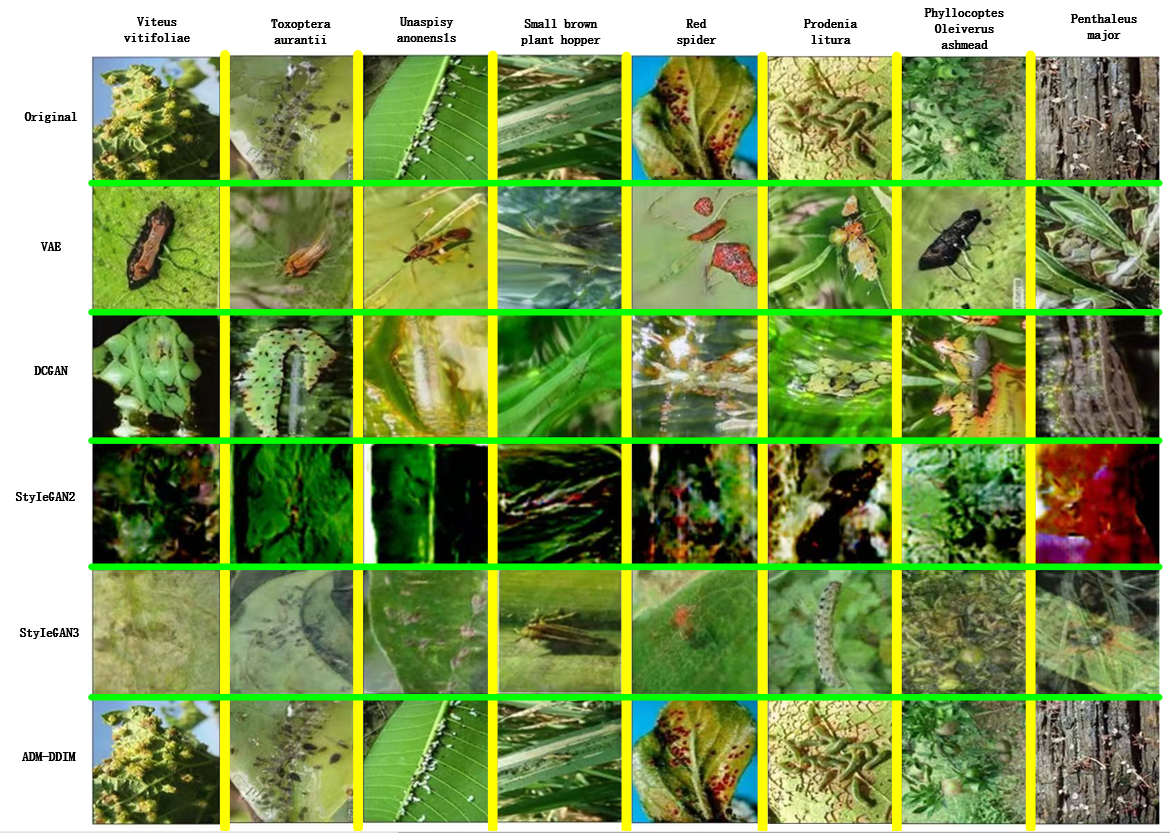} \hfil \caption{Real samples and generated samples in the IP102 pest dataset. The real samples are from the training set, and the generated samples are generated by ADM-DDIM, VAE and GAN. Each row corresponds to a kind of pest. \cite{96}} \label{image13} \end{figure}

To evaluate the image generation effect of different generative models, this paper conducts experiments on the large-scale IP102 pest dataset, and compares five methods: VAE \cite{114}, DCGAN \cite{120}, StyleGAN2 \cite{121}, StyleGAN3 \cite{122} and the ADM-DDIM model proposed in this paper. The results are shown in Figure~\ref{image13}. ADM-DDIM performs best in generating realistic and detailed pest images, especially in texture synthesis and detail retention. It still has stable performance even in the case of strong category diversity and serious category imbalance.

Table~\ref{tab2} shows that ADM-DDIM performs well in all indicators, especially in FID reaching 23.66, IS reaching 7.26, and NIQE reaching 3.05, which shows its high-quality generation ability in agricultural pest image generation.

In contrast, the traditional generative model VAE performed poorly in capturing image details and reconstructing complex pest textures, with FID as high as 264.57, IS as low as 2.39, and NIQE reaching 7.52.

Although DCGAN, StyleGAN2, and StyleGAN3 can generate high-resolution images with good details, their performance is still limited when dealing with high diversity and severely unbalanced data. For example, StyleGAN3 has an FID of 42.54, an IS of 4.70, and a NIQE of 4.08, which is worse than ADM-DDIM.

These results show that ADM-DDIM has obvious advantages in generating diverse and high-fidelity pest images. At the same time, it effectively alleviates the problems of large intra-class differences and serious inter-class imbalance in agricultural image datasets, showing good application potential.


\begin{table}[htbp]
\centering
\caption{Performance comparison between diffusion models and other generative models.\cite{96}}
\label{tab2}
\renewcommand{\arraystretch}{1.4}
\small   
\begin{tabular}{c c c c c c c c}
\toprule
\multirow{2}{*}{\textbf{Models}} & \multirow{2}{*}{\textbf{Original}} & \multicolumn{2}{c}{\textbf{VAE}} & \multicolumn{2}{c}{\textbf{GANs}} & & \textbf{Diffusion Model} \\
\cmidrule(lr){3-4} \cmidrule(lr){5-6} \cmidrule(lr){8-8}
& & VAE & DCGAN & StyleGAN2 & StyleGAN3 & & ADM-DDIM \\
\midrule
FID$\downarrow$     & /    & 264.57 & 217.34 & 86.90 & 42.52 & & 23.66 \\
IS$\uparrow$        & 1.40 & 2.39   & 2.58   & 2.70  & 4.70  & & 7.26 \\
NIQE$\downarrow$    & 4.22 & 7.52   & 7.71   & 6.28  & 4.08  & & 3.05 \\
\bottomrule
\end{tabular}
\end{table}

\begin{table}[htbp]
\centering
\caption{Validation accuracy (val-acc \%) of different models. The maximum value indicates the maximum accuracy improvement.\cite{96}}
\label{tab3}
\renewcommand{\arraystretch}{1.4}
\setlength{\tabcolsep}{4pt}
\resizebox{0.8\textwidth}{!}{
\begin{tabularx}{\textwidth}{C C C C C C C C}
\toprule
& \textbf{Models} & \textbf{ResNet} & \textbf{GoogLeNet} & \textbf{ShuffleNet} & \textbf{DenseNet} & \textbf{EfficientNet} & \textbf{MobileNet} \\
\midrule
\textbf{Original}   &             & 69.10 & 63.60 & 65.30 & 68.90 & 70.30 & 62.50 \\
\multirow{5}{*}{Generated}
& VAE         & 68.70 & 64.50 & 64.10 & 67.60 & 70.10 & 63.70 \\
& DCGAN       & 69.80 & 64.60 & 64.90 & 69.10 & 71.00 & 64.40 \\
& StyleGAN2   & 69.60 & 65.20 & 65.40 & 69.00 & 71.40 & 63.40 \\
& StyleGAN3   & 69.20 & 65.70 & 65.60 & 69.40 & 70.60 & 63.80 \\
& ADM-DDIM    & 73.80 & 69.30 & 70.70 & 74.00 & 74.80 & 69.10 \\
\midrule
\textbf{Maximum$\uparrow$} & & \textbf{4.70}$\uparrow$ & \textbf{5.70}$\uparrow$ & \textbf{5.40}$\uparrow$ & \textbf{5.10}$\uparrow$ & \textbf{4.50}$\uparrow$ & \textbf{6.60}$\uparrow$ \\
\bottomrule
\end{tabularx}}
\end{table}

Table~\ref{tab3} shows the classification results of the images generated by the ADM-DDIM model under six classification architectures.
For ResNet, the validation accuracy on the StyleGAN2-augmented dataset was 69.60\%, only 0.50\% higher than the baseline.  
With ADM-DDIM, the accuracy of ResNet increased to 73.80\%, which is 4.2\% higher than StyleGAN2 and 4.7\% higher than the baseline.  

EfficientNetV2 gained almost no benefit from StyleGAN2, with accuracy improving by just 1.1\%.  
By contrast, ADM-DDIM lifted its accuracy to 74.80\%, which represents a 4.5\% increase over the dataset without augmentation.

The same tendency appeared in other models.  
GoogLeNet recorded 69.30\%, improving on StyleGAN2 by 4.1\% and on StyleGAN3 by 3.6\%.  
MobileNet showed the strongest response, jumping from 62.50\% to 69.10\%, a gain of 6.6\%.  

Overall, these comparisons suggest that ADM-DDIM is effective in reducing the impact of long-tail data and in raising the classification performance of deep learning models used for agricultural tasks.  
In contrast, datasets generated by DCGAN or VAE reduced accuracy because the synthetic images contained artifacts or distortions, which hindered the training and generalization of the classifiers.  

\subsubsection{Remote Sensing and Hyperspectral Image Reconstruction Experiment}

Table~\ref{tab4} compares the performance of several models on the CropSR-Test dataset.  
To test the proposed enhanced variance-aware diffusion model (EVADM) with variance-aware spatial attention (VASA), we evaluated it against the regression-based EDSR \cite{132}, the GAN-based Real-ESRGAN \cite{124}, and the latent diffusion model (LDM) \cite{59}.

The experimental results show that the introduction of VASA attention mechanism significantly improves the model performance with only a small increase in computational complexity.
At the $\times2$ SR scale, the VASR-fc1-x2 variant achieves the best performance, primarily due to its better structural similarity index.

At the $\times4$ SR multiple, EVADM-x4 is superior to LDM-x4 in both FID and SRFI indicators, with FID decreasing by 5.7\% and SRFI increasing by 7\%.
In addition, the EVADM model reduces the computational complexity by more than 10\% and the number of parameters by nearly 50\%, highlighting its high efficiency and practicality in high-resolution agricultural image reconstruction.

\begin{table}[htbp]
\centering
\caption{Comparison of model performance on CropSR-Test.\cite{95}}
\label{tab4}
\renewcommand{\arraystretch}{1.2}
\setlength{\tabcolsep}{2pt}
\resizebox{0.8\textwidth}{!}{
\begin{tabular}{lccccccccc}
\hline
\textbf{Model} & \textbf{Type} & \textbf{r} & \textbf{PSNR$\uparrow$} & \textbf{SSIM$\uparrow$} & \textbf{FID$\downarrow$} & \textbf{RFID*$\uparrow$} & \textbf{SRFI$\uparrow$} & \textbf{FLOPs/G$\downarrow$} & \textbf{Parms/M$\downarrow$} \\
\hline
EDSR\_x2           & Reg  & 2  & 26.18 & 0.90 & 6.07  & 0.70 & 14.17 & 90.18   & 1.37 \\
\textbf{VASR\_fc1\_x2} & Reg  & 2  & \textbf{26.63} & \textbf{0.92} & \textbf{4.74} & \textbf{0.77} & \textbf{16.74} & 90.87   & 1.45 \\
RealESRGAN\_x2     & GAN  & 2  & 21.75 & 0.78 & 27.13 & -0.34 & 1.11  & 294.24  & 37.78 \\
EVADM\_x2          & Diff & 2  & 22.69 & 0.83 & 10.01 & 0.50 & 8.29  & 22.97   & 29.14 \\
EDSR\_x4           & Reg  & 4  & \textbf{19.18} & \textbf{0.49} & 51.78 & 0.53 & 2.44  & 130.29  & 1.52 \\
RealESRGAN\_x4     & GAN  & 4  & 17.27 & 0.44 & 40.38 & 0.63 & 2.67  & 1176.61 & 37.77 \\
VARDGAN\_x4        & GAN  & 4  & 17.43 & 0.45 & 31.73 & 0.71 & 2.95  & 1188.19 & 40.71 \\
LDM\_x4            & Diff & 4  & 16.37 & 0.35 & 31.36 & 0.71 & 2.79  & 40.22   & 113.62 \\
EVADM\_ca\_x4      & Diff & 4  & 16.21 & 0.34 & 25.98 & 0.76 & 2.93  & 35.96   & 64.03 \\
\textbf{EVADM\_x4} & Diff & 4  & 16.49 & 0.36 & \textbf{25.66} & \textbf{0.77} & \textbf{2.97} & 35.91   & 63.20 \\
EDSR\_x8           & Reg  & 8  & \textbf{16.79} & \textbf{0.20} & 131.24 & 0.41 & 1.34  & 290.60  & 1.67 \\
RealESRGAN\_x8     & GAN  & 8  & 14.84 & 0.13 & 280.12 & -0.27 & 0.89  & 1453.58 & 37.77 \\
\textbf{EVADM\_x8} & Diff & 8  & 14.30 & 0.14 & \textbf{64.14} & \textbf{0.71} & \textbf{1.57} & 35.92   & 63.21 \\
\hline
\end{tabular}
}
\vspace{0.3em}
\end{table}

\begin{figure}[htbp] \centering \includegraphics[width=0.7\textwidth]{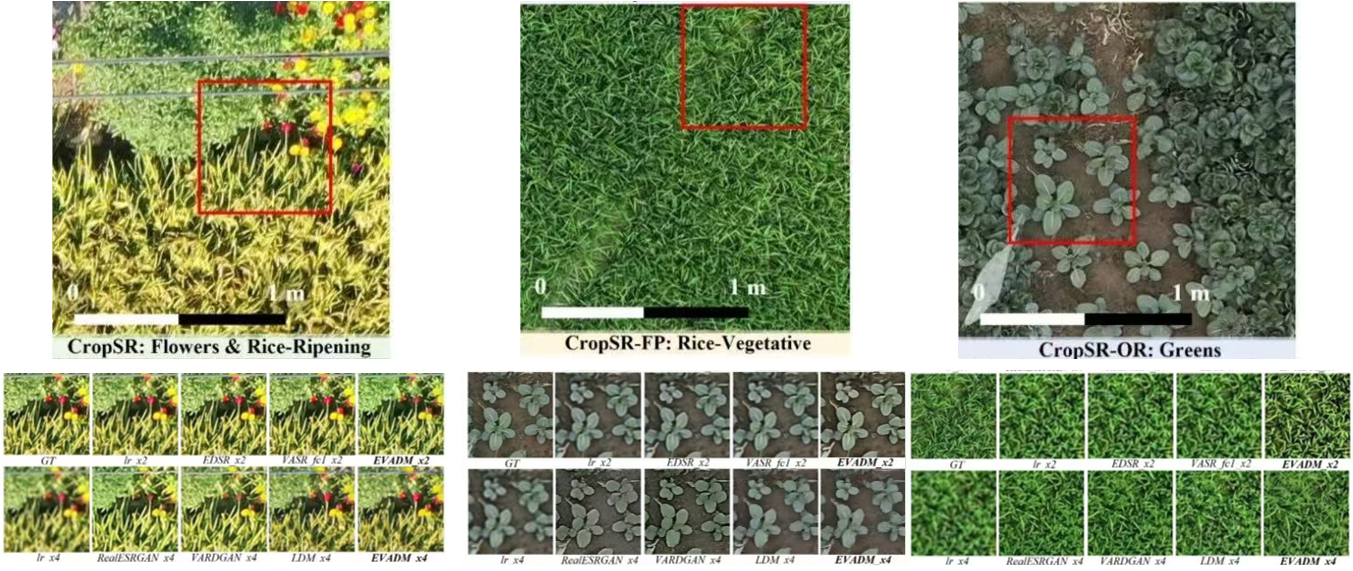} \hfil \caption{Visual comparison of super-resolution results on CropSR-Test, CropSR-OR, and CropSR-FP datasets. The left side shows an overview of different farmland scenes, and the model name is marked under the small tiles on the right. \cite{95}} \label{image14} \end{figure}

Figure~\ref{image14} shows the visual comparison results of representative models on the CropSR-Test dataset.

At the $\times2$ super-resolution (SR) scale, the images generated by VASR-fc1-x2 present rich details, bright colors, and clear edges. Although EVADM-x2 is more realistic in reconstructing the crop canopy structure, it is relatively insufficient in dealing with rare flower categories.

In the more challenging $\times4$ SR task, images generated by GAN-based models showed significant distortion. In contrast, diffusion models, while failing to achieve a perfect reconstruction of the ground truth, are able to reproduce similar structural patterns and perform better in terms of visual fidelity.

\begin{figure}[htbp] \centering \includegraphics[width=0.7\textwidth]{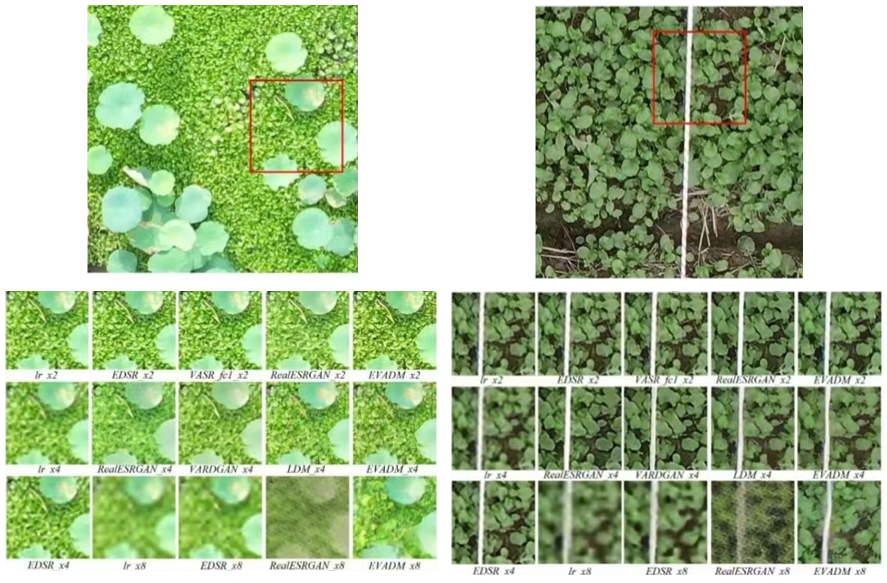} \hfil \caption{Visual comparison of super-resolution results on the CropSR-Test dataset. The left side shows the overview of different farmland scenes, and the model name is marked below the right-side image tiles. \cite{95}} \label{image15} \end{figure}

In the case of $\times8$ SR enhancement, the model can only infer the general features due to the serious loss of information. Nevertheless, compared with the regression model and the GAN model, the results generated by EVADM-x8 (Figure~\ref{image15}) have clearer edges and more distinguishable details. It is worth noting that its FID score is reduced by 67.1, which further verifies the effectiveness and feasibility of the diffusion model in the agricultural image extreme scale super-resolution task.

The test results on the CropSR-OR/FP dataset are shown in Table~\ref{tab5}. Although the two regression models EDSR-x2 and VASR-fc1-x2  achieve higher scores in the SSIM metric, their super-resolution results are noticeably blurry, close to the visual effect of simple interpolation, as shown in the middle (CropSR-OR) and bottom (CropSR-FP) of Figure~\ref{image15}.

In both $\times2$ and $\times4$ super-resolution tasks, the proposed EVADM model achieves the best performance metrics in terms of FID and SRFI.Specifically, EVADM-x2 reduces the FID score by 14.6 on average, while EVADM-x4 reduces it by 8.0, compared with the baseline model. Correspondingly, SRFI increased by 27.1\% and 6.3\% respectively.
By comparison, the GAN-based models produced RFID scores that were nearly zero, which shows their limited ability to deliver acceptable super-resolution results under the specified evaluation settings.  
As can be seen from the VARDGAN-x4 results of the second sample in Figure~\ref{image15}, the model fails to preserve the detailed features of the leaves, while EVADM-x4 can reconstruct the leaf veins more clearly. In the $\times4$ super-resolution task of the dense crop canopy in the third row, the GAN model fail to effectively restore narrow leaves.

Comparing the last two images generated by the $\times4$ diffusion model, we found that the introduction of the VASA attention mechanism significantly improves the model's ability to reconstruct detailed structures. In addition, the performance of the regression model on the simulated CropSR-Test dataset was compared with its performance on the real dataset, it can be observed that the numerical indicators and visual quality are significantly reduced, which indicates that its generalization ability is limited. In contrast, the diffusion model exhibits stronger robustness in the cross-domain case, further confirming its potential in real agricultural applications.

\begin{table}[htbp]
\centering
\caption{Comparison of CropSR-OR and CropSR-FP model performance.\cite{95}}
\label{tab5}
\renewcommand{\arraystretch}{1.2}
\setlength{\tabcolsep}{4pt}
\begin{tabular}{cccccccccc}
\hline
\multirow{2}{*}{\textbf{Model}} & \multirow{2}{*}{\textbf{r}} 
& \multicolumn{4}{c}{\textbf{CropSR-OR}} 
& \multicolumn{4}{c}{\textbf{CropSR-FP}} \\ 
& & SSIM$\uparrow$ & FID$\downarrow$ & RFID$^*$ $\uparrow$ & SRFI$\uparrow$ & SSIM$\uparrow$ & FID$\downarrow$ & RFID$^*$ $\uparrow$ & SRFI$\uparrow$ \\ 
\hline
EDSR\_x2           & 2 & 0.55 & 66.60 & 0.13 & 2.54 & 0.50 & 78.71 & 0.14 & 2.46 \\
VASR\_fc1\_x2      & 2 & 0.55 & 65.96 & 0.14 & 2.60 & 0.50 & 78.14 & 0.15 & 2.49 \\
\textbf{EVADM\_x2} & 2 & 0.42 & \textbf{52.30} & 0.32 & \textbf{3.36} & 0.35 & \textbf{62.50} & 0.32 & \textbf{3.11} \\ RealESRGAN\_x4     & 4  & 0.29 & 124.75 & 0.00 & 1.17 & 0.35 & 141.89 & 0.11 & 1.39 \\
VARDGAN\_x4        & 4 & 0.30 & 116.09 & 0.07 & 1.28 & 0.34 & 126.24 & 0.21 & 1.55 \\
LDM\_x4            & 4 & 0.28 & 61.75 & 0.50 & 2.10 & 0.35 & 92.47 & 0.42 & 1.98 \\
\textbf{EVADM\_x4} & 4 & 0.28 & \textbf{52.47} & \textbf{0.58} & 2.28 & 0.33 & \textbf{85.78} & 0.46 & \textbf{2.06} \\
EVADM\_x8          & 8 & 0.15 & 101.72 & 0.57 & 1.45 & 0.21 & 151.09 & 0.39 & 1.33 \\
\hline
\end{tabular}
\end{table}

\subsubsection{Experiment on Diagnosis and Auxiliary Decision-making of Plant Diseases and Insect Pests}

In this experiment, the performance evaluation of sunflower disease detection was carried out, and the indicators used included precision, recall, accuracy, F1 value and mean Average Precision (mAP@75). The results show that the combination of structural optimization and task customization can significantly improve the model effect, especially in the complex background and small sample conditions.

As shown in Table~\ref{tab6}, RT-DETRv2 has the lowest overall performance. It relies on global feature extraction, which makes it difficult to capture fine-grained lesion information. Mask R-CNN performs better because it has the ability of instance segmentation, but it still has some shortcomings in the segmentation of small lesions. The results of the DETR model are relatively reasonable, but there are some difficulties in suppressing background interference and identifying small-scale lesions.

Among the compared approaches, the LeafDetection model \cite{125}, developed specifically for sunflower disease detection has shown significant advantages. With strong capability in capturing fine-grained features, it outperforms all other models on every evaluation metric, reaching an accuracy of 0.94, recall of 0.92, precision of 0.93, F1 score of 0.93, and mAP@75 of 0.92. These results highlight the effectiveness of combining deep learning methods with domain-oriented feature modeling.  

On this basis, the authors put forward a new framework called the \textbf{Few-Shot Diffusion Detection Network (FSDDN)}. This approach integrates few-shot learning with diffusion models, enabling accurate and adaptable disease detection under data-scarce conditions, and is therefore well suited to agricultural environments where labeled samples are limited.  

\vspace{0.5em}
Apart from the experiments conducted on the original dataset,The robustness and generalization capability of the proposed method were also evaluated using a separate dataset. This additional set contains four categories of leaf images: downy mildew (120 images), fresh leaves (134 images), gray mold (72 images), and leaf scars (140 images).  
Its diversity is reflected in the large differences in the types of diseases and leaf morphology. Among them, the boundaries of downy mildew and gray mold lesions are blurred, and they are easily disturbed by the background.

The annotation procedure for this dataset adheres to the same standards as the original one, and the precise annotation box of the diseased area is provided by experts. In order to alleviate the category imbalance, data equalization and enhancement strategies are introduced in the training, including rotation, cropping and color disturbance. These operations not only improve the robustness of the model, but also effectively expand the training samples.
\begin{table}[htbp]
\centering
\caption{Experimental results of disease detection model.\cite{107}}
\label{tab6}
\renewcommand{\arraystretch}{1.2}
\resizebox{0.8\textwidth}{!}{
\begin{tabularx}{\textwidth}{lXXXXX}
\hline
\textbf{Model} & \textbf{Precision} & \textbf{Recall} & \textbf{Accuracy} & \textbf{mAP@75} & \textbf{F1-Score} \\
\hline
RT-DETRv2       & 0.83 & 0.80 & 0.82 & 0.81 & 0.81 \\
Mask R-CNN      & 0.87 & 0.84 & 0.86 & 0.85 & 0.85 \\
DETR            & 0.89 & 0.86 & 0.87 & 0.87 & 0.87 \\
LeafDetection   & 0.92 & 0.89 & 0.91 & 0.90 & 0.90 \\
FSDDN           & 0.94 & 0.92 & 0.93 & 0.92 & 0.93 \\
\hline
\end{tabularx}}
\end{table}

Table~\ref{tab7} makes it clear that the integration of these strategies produced notable improvements in model performance, underscoring both the robustness and the broad adaptability of the proposed approach when applied to complex and imbalanced agricultural datasets.  

\begin{figure}[htbp]
\centering
\includegraphics[width=0.7\textwidth]{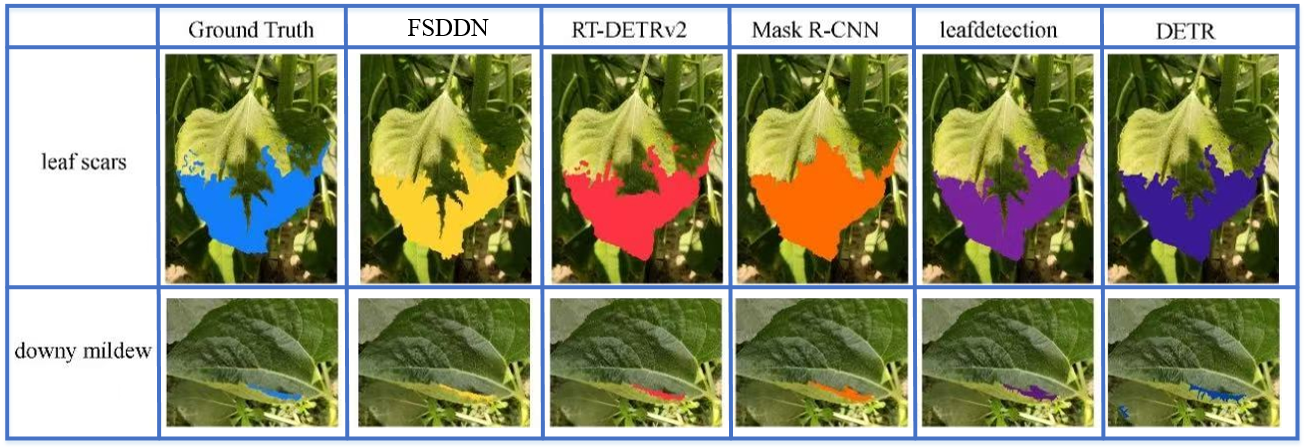}%
\hfil
\caption{Visualization analysis of experimental results.\cite{107}}
\label{image16}
\end{figure}

\vspace{0.5em}
Figure~\ref{image16} presents a comparative visualization of the results produced by RT-DETRv2, Mask R-CNN, DETR, LeafDetection, and the proposed approach.  
RT-DETRv2 mainly relies on global feature extraction, so it is insufficient in detail capture. Mask R-CNN performs well in instance segmentation, but it lacks accuracy in small-scale lesions. DETR relies on the global attention mechanism, but it still has shortcomings in detecting smaller lesions. LeafDetection has a more balanced overall performance and has certain advantages in detail feature extraction.

In contrast, the proposed method is better than other models in all disease categories. In particular, when dealing with complex diseases such as downy mildew and leaf scars, its segmentation and recognition are more reliable, showing higher robustness and stability.


\begin{table}[htbp]
\centering
\caption{Experimental results of disease detection models on other datasets.\cite{107}}
\label{tab7}
\renewcommand{\arraystretch}{1.2}
\resizebox{0.8\textwidth}{!}{
\begin{tabularx}{\textwidth}{l*{5}{>{\centering\arraybackslash}X}}
\hline
\textbf{Model} & \textbf{Precision} & \textbf{Recall} & \textbf{Accuracy} & \textbf{mAP@75} & \textbf{F1-Score} \\
\hline
RT-DETRv2       & 0.84 & 0.80 & 0.82 & 0.81 & 0.82 \\
Mask R-CNN      & 0.86 & 0.82 & 0.84 & 0.83 & 0.84 \\
DETR            & 0.87 & 0.85 & 0.86 & 0.86 & 0.86 \\
LeafDetection   & 0.90 & 0.89 & 0.90 & 0.90 & 0.89 \\
FSDDN           & 0.92 & 0.90 & 0.91 & 0.91 & 0.91 \\
\hline
\end{tabularx}}
\end{table}

\subsubsection{Multimodal Fusion Experiment}

In this experiment, a deep learning model is proposed for detecting jujube tree diseases in desert environments which is basing on multimodal data fusion. Because of the intense sunlight and extreme environmental variations in desert areas, the existing methods have some shortcomings in feature extraction and detection accuracy. To solve these problems, this study combines image data with sensor information, and introduces a feature extraction mechanism that integrates Transformer and diffusion modules has been introduced to achieve accurate capture of disease features.

\begin{figure}[htbp]
\centering
\includegraphics[width=0.7\textwidth]{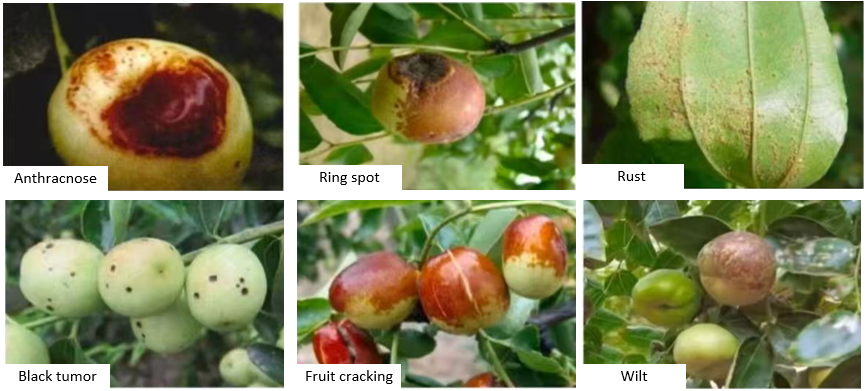}%
\hfil
\caption{Samples of different jujube tree diseases: (a) jujube wilt disease, (b) fruit cracking disease, (c) black tumor disease, (d) rust disease, (e) ring spot disease, (f) anthracnose disease.\cite{112}}
\label{image17}
\end{figure}

\begin{figure}[htbp]
\centering
\includegraphics[width=0.7\textwidth]{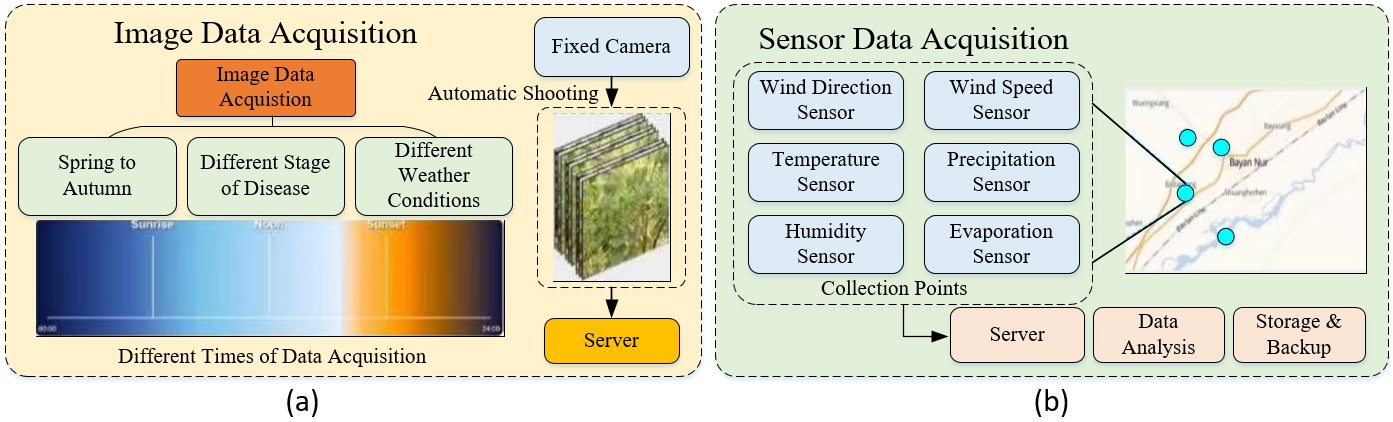}%
\hfil
\caption{Overview of data collection. (a) Image data collection: images are acquired at different times, seasons, disease stages and conditions, and stored in the central server. (b) Sensor data acquisition: Environmental data (such as wind speed, temperature, and humidity) are collected through a variety of sensors and transmitted for analysis and storage. \cite{112}} \label{image18} \end{figure}

The data collection was mainly conducted in the Bayan Nur desert area in northern China, where environmental factors  have a significant impact on jujube tree diseases such as light, temperature and humidity. To obtain disease information under different environmental conditions, data collection covered the entire cycle from spring to autumn, encompassing various growth stages of jujube trees and the development process of diseases.Images were taken in the early morning, noon and evening every day (as shown in Figure~\ref{image17} and Figure~\ref{image18}(a)) to ensure that the model can identify diseases under different lighting conditions, thereby enhancing the generalization ability of the model and avoiding overfitting. At the same time, field sensors function nonstop to record key environmental parameters, including temperature, humidity, and wind speed.  
 By integrating these environmental measurements, the model gains deeper insight into how external conditions drive disease development, which in turn enhances its overall resilience and stability.

Beyond visual imagery, environmental variables—including temperature, humidity, and wind speed—were tracked in real time using field sensors. By merging this contextual data with the image inputs, the model can uncover hidden associations between disease progression and microclimatic fluctuations, thereby enhancing its capacity to adapt to real-world field conditions.  

Table~\ref{tab8} compares the performance of YOLOv9 \cite{126}, Retinanet \cite{127}, Efficientdet \cite{128}, YOLOv10 \cite{129} and DETR \cite{130} for disease identification tasks, and evaluates their convergence speed in the training process and the final detection results. Table~\ref{tab9} compares the effectiveness of the proposed model against representative state-of-the-art approaches for disease identification.

\begin{table}[htbp]
\centering
\caption{Comparison with different mainstream models\cite{112}}
\label{tab8}
\renewcommand{\arraystretch}{1.2}
\scriptsize
\resizebox{0.8\textwidth}{!}{
\begin{tabularx}{\textwidth}{l*{6}{>{\centering\arraybackslash}X}}
\hline
\textbf{Metric} & \textbf{YOLOv9} & \textbf{Retinanet} & \textbf{Efficientdet} & \textbf{YOLOv10} & \textbf{DETR} & \textbf{JuDiformer} \\
\hline
Precision      & 0.83   & 0.85   & 0.87   & 0.89   & 0.91   & 0.93 \\
Recall         & 0.79   & 0.81   & 0.83   & 0.85   & 0.87   & 0.89 \\
Accuracy       & 0.81   & 0.83   & 0.85   & 0.86   & 0.88   & 0.90 \\
mAP            & 0.82   & 0.84   & 0.86   & 0.87   & 0.89   & 0.91 \\
F1 Score       & 0.81   & 0.83   & 0.85   & 0.87   & 0.89   & 0.91 \\
FPS            & 35     & 25     & 30     & 40     & 22     & 28   \\
Params (M)     & 57.3   & 52     & 77     & 29.5   & 60     & 61.32 \\
FLOPs (G)      & 189    & 198    & 410    & 160    & 253    & 207  \\
Memory (GB)    & 10     & 8      & 12     & 9      & 11     & 10   \\
\hline
\end{tabularx}}
\end{table}

\begin{table}[htbp]
\centering
\caption{Comparison with other state-of-the-art results.\cite{112}}
\label{tab9}
\renewcommand{\arraystretch}{1.2}
\begin{tabularx}{\textwidth}{l*{5}{>{\centering\arraybackslash}X}}
\hline
\textbf{Model} & \textbf{Precision} & \textbf{Recall} & \textbf{Accuracy} & \textbf{mAP} & \textbf{F1 Score} \\
\hline
Thai et al~\cite{131}        & 0.76 & 0.71 & 0.73 & 0.72 & 0.73 \\
Zeng et al~\cite{132}        & 0.78 & 0.73 & 0.75 & 0.74 & 0.75 \\
Pang et al~\cite{133}        & 0.79 & 0.74 & 0.76 & 0.75 & 0.77 \\
Li et al~\cite{134}          & 0.81 & 0.76 & 0.78 & 0.77 & 0.78 \\
Sebastian et al~\cite{135}   & 0.82 & 0.78 & 0.79 & 0.79 & 0.80 \\
Wang et al~\cite{136}        & 0.84 & 0.79 & 0.81 & 0.81 & 0.81 \\
Zhou et al~\cite{137}        & 0.85 & 0.81 & 0.82 & 0.82 & 0.83 \\
Sharma et al~\cite{138}      & 0.87 & 0.82 & 0.84 & 0.84 & 0.85 \\
Wang et al~\cite{139}        & 0.88 & 0.84 & 0.85 & 0.86 & 0.86 \\
Karthik et al~\cite{140}     & 0.90 & 0.86 & 0.87 & 0.88 & 0.88 \\
Zhang and Lu et al~\cite{141}& 0.91 & 0.87 & 0.88 & 0.89 & 0.89 \\
JuDifformer                  & 0.93 & 0.89 & 0.90 & 0.91 & 0.91 \\
\hline
\end{tabularx}
\end{table}

Table~\ref{tab10} shows that when only image data is used, the model's indicators are as follows: Precision is 0.81, Recall is 0.77, Accuracy is 0.79, mAP is 0.78, and F1-score is 0.79; when only sensor data is used, these indicators are increased to 0.85, 0.82, 0.83, 0.83 and 0.83 respectively, indicating that the environmental information in the sensor data is closely related to the disease process. Furthermore, when the image and sensor data are fused, all indicators are significantly improved: Precision reaches 0.93, Recall is 0.89, Accuracy is 0.90, mAP is 0.91, and F1-score is 0.91, indicating that multi-modal data fusion has a significant effect on improving the accuracy and robustness of disease detection.

\begin{table}[htbp]
\centering
\caption{Comparison with different multimodal datasets.\cite{112}}
\label{tab10}
\renewcommand{\arraystretch}{1.2}
\begin{tabularx}{\textwidth}{l*{5}{>{\centering\arraybackslash}X}}
\hline
\textbf{Data} & \textbf{Precision} & \textbf{Recall} & \textbf{Accuracy} & \textbf{mAP} & \textbf{F1 Score} \\
\hline
Image   & 0.81 & 0.77 & 0.79 & 0.78 & 0.79 \\
Sensor  & 0.85 & 0.82 & 0.83 & 0.83 & 0.83 \\
Both    & 0.93 & 0.89 & 0.90 & 0.91 & 0.91 \\
\hline
\end{tabularx}
\end{table}

\subsection{Key Findings and Conclusions of the Experiment}

The outcomes of the four experimental tasks collectively emphasize the strong potential of diffusion models in agricultural scenarios, with particular advantages in pest and disease detection as well as data augmentation.

For data augmentation and image generation, the diffusion-based approach (ADM-DDIM) markedly enhanced the realism and quality of synthetic images. Compared with generative baselines like VAE, DCGAN, StyleGAN2, and StyleGAN3, the ADM-DDIM approach delivered clear improvements on evaluation indices such as FID, IS, and NIQE. It generated outputs with more realistic texture reproduction and finer structural detail, while at the same time alleviating the challenges posed by imbalanced long-tail distributions. Another noteworthy advantage was its enhanced capability to recognize rarely occurring pest categories.
Taken together, these advances illustrate how diffusion-based techniques can serve as an effective solution to the enduring problems of data scarcity and category imbalance in agricultural image datasets.

In the context of remote sensing and hyperspectral image reconstruction, the EVADM model, augmented with the variance-aware spatial attention mechanism (VASA), delivered strong results in super-resolution tasks. When compared with regression-driven methods (e.g., EDSR) and GAN-based techniques (e.g., RealESRGAN, LDM), EVADM recorded significant gains in FID and SRFI, confirming both its reliability and effectiveness for agricultural image enhancement.In addition to achieving strong accuracy, EVADM reduced computational overhead and adopted a lighter architecture, which makes it more practical for deployment in real agricultural environments.

For pest and disease detection, generative enhancement with diffusion models provided significant gains in both accuracy and robustness, particularly in scenarios characterized by long-tail distributions or scarce training samples. Through the creation of realistic synthetic samples, diffusion models enhanced the detection of seldom-seen categories, exemplified by sunflower diseases and date palm diseases occurring in arid environments.

Moreover, the integration of multimodal inputs (e.g., combining imagery with sensor-derived data) further increased detection precision and adaptability in challenging agricultural contexts.

Taken together, these results demonstrate that diffusion models are powerful tools for data augmentation, image reconstruction, and multimodal integration. They offer reliable and precise support for smart agriculture, especially in pest and disease surveillance, while contributing to accelerated innovation and technological growth across the agricultural domain.

\section{Challenges and Countermeasures of Diffusion Model in Agricultural Application}\label{sec6}

Although diffusion models have shown great promise for agricultural advancement—ranging from pest detection to crop monitoring—several key challenges remain before their full potential can be realized.

\subsection{Challenges}

\begin{itemize}
    \item \textbf{Computational efficiency:} Generating high-resolution images with diffusion models demands intensive computation, a limitation that becomes more acute when operating on large-scale datasets. Their step-by-step sampling procedure, which requires numerous network passes, leads to slow inference speeds. Such delays pose a serious challenge for real-time agricultural applications, where timely outputs are essential. To overcome this, future studies should emphasize the design of more efficient sampling algorithms and make greater use of distributed or parallel computing to reduce latency and enhance scalability.

\item \textbf{Data scarcity and imbalance:} Agricultural datasets are frequently incomplete, especially for rare or newly emerging pests and diseases. In the initial stages of an outbreak, the limited availability of samples makes it difficult to train reliable models. At the same time, the dominance of certain crops or diseases in datasets can distort learning and weaken generalization. A practical way forward is to integrate powerful augmentation pipelines with the generation of trustworthy synthetic examples of high fidelity. By applying these strategies, the negative impact of class imbalance can be alleviated, and models gain enhanced robustness, allowing them to deliver consistent performance even when training data are limited.  

\item \textbf{Generalization across diverse environments:} The wide variation in agricultural conditions—shaped by changes in climate, differences in soil properties, and geographical diversity—creates substantial difficulties for ensuring that models generalize effectively across settings.
Diffusion models frequently struggle when faced with new or less-represented contexts, leading to degraded accuracy outside controlled conditions. Enhancing generalization requires exploring techniques such as domain adaptation, transfer learning, and multimodal fusion so that models can adapt effectively across wide-ranging agricultural environments.

    \item \textbf{Dataset quality:} The reliability of diffusion-based methods is highly sensitive to the training data used. Low-resolution images, labeling errors, and limited variability in environmental conditions can severely hinder model training and reduce the realism of generated outputs. To guarantee reliable and applicable results in agricultural practice, it is essential to establish datasets that are extensive, well-annotated, and rich in diversity.  
\end{itemize}

\subsection{Future Research Directions}

Looking forward, multiple directions could further strengthen how diffusion models serve agricultural needs.

\begin{itemize}
    \item \textbf{Raising computational efficiency:} The heavy resource demands of diffusion models limit their speed in practice. Future studies should aim at designing compact network structures, creating faster sampling techniques, and refining implementation strategies. Making use of modern hardware accelerators will also be important to cut inference delays without losing image fidelity. Recent surveys have systematically summarized efficient diffusion methods from the algorithmic and system levels~\cite{liu2025efficient}, and speculative sampling has been demonstrated as a promising solution to accelerate inference~\cite{shen2025speculative}.
    
    \item \textbf{Generating stronger synthetic data:} In agriculture, datasets are frequently limited in size and skewed in distribution. Addressing this requires advanced data synthesis methods that can enlarge training sets, balance class representation, and improve model performance across diverse scenarios. New methods are needed to generate realistic synthetic samples. This can expand the dataset, improve the representation of rare categories, and enhance the stability and accuracy of the model in a variety of crops, diseases and pests. For instance, diffusion-based generative augmentation has been shown to improve plant disease classification~\cite{93}, and iterative synthetic dataset construction has proven effective for tomato disease detection~\cite{lin2024synthetic}.
    
    \item \textbf{Enhancing cross-domain generalization:} Agricultural conditions differ widely, so models must adapt effectively across domains. Techniques such as domain adaptation and transfer learning offer practical solutions. In addition, integrating multimodal information—including imagery, sensor data, and climate records—can lead to stronger, context-aware agricultural systems. Recent reviews highlight that generative models combined with computer vision are key to improving cross-domain generalization in agriculture~\cite{gao2025cv}.
    
    \item \textbf{Integrating with other generative frameworks:} Bringing diffusion models together with approaches like GANs or VAEs can create generative pipelines that are both more flexible and more efficient. Hybrid schemes allow finer control over the synthesis process and point to fresh opportunities in agricultural research. Future investigations should also consider semantic-level representations, which would enable targeted generation of agricultural data, such as images of particular crop diseases or growth stages. Recent advances such as latent denoising diffusion GANs demonstrate that hybrid methods can simultaneously achieve faster sampling and higher image quality~\cite{kim2024lddgan}.
\end{itemize}

By tackling these challenges and advancing along these directions, diffusion models are poised to become valuable assets in modern agriculture, especially for pest and disease detection, crop monitoring, and precision farming.

\section{Summary}\label{sec7}

Diffusion-based methods demonstrate significant promise for agricultural use cases. They are particularly well suited for crop health assessment, pest and disease surveillance, and the analysis of remote sensing imagery. Their ability to create high-quality synthetic data makes them an effective tool to overcome common obstacles in agriculture, including limited data availability and imbalanced class distributions.
By generating high-quality synthetic data, diffusion models effectively alleviate common problems in agriculture, such as insufficient data and category imbalance. In the tasks of fine-grained disease identification and crop growth prediction, these models can provide clearer and more detailed visualization results of crop health, and provide support for the scientific decision-making of agricultural experts.

Different from conventional approaches, diffusion models create images through a step-by-step denoising process. This progressive strategy brings out richer details and higher-quality results. They are especially valuable for data augmentation and image generation, where realistic synthetic samples can be produced to enlarge limited datasets, increase training material, and strengthen the performance of downstream recognition tasks.

In practical scenarios such as pest detection, disease identification, or crop growth prediction, the images produced by diffusion models show both high fidelity and strong adaptability to complex environments. Such flexibility enhances model robustness and raises prediction accuracy. With conditional generation, these models can further produce outputs tailored to specific crop varieties, disease cases, or environmental factors, thereby meeting the diverse needs of precision agriculture.

A key strength of diffusion models is their capability for multimodal integration. They can bring together heterogeneous data—such as images, sensor readings, and climate information—into a unified structure. This ability expands their practical value, supporting more accurate pest and disease monitoring, deeper analysis of environmental variations, and more reliable decision-making in precision agriculture.


\section*{Declarations}
\textbf{Funding} This work is jointly supported by the Shanghai Agricultural Technology Innovation Project (2024-02-08-00-12-F00032), the High-Quality Development Special Project of the Ministry of Industry and Information Technology (TC240A9ED-56), and the Academician Workstation Program of Yunnan Province (202405AF140013).
\section*{Author Contributions}

Xing Hu: Conceptualization, Literature Review, Methodology Design, Manuscript Writing, and Project Supervision. 
Haodong Chen: Data Collection, Figure and Table Preparation, and Draft Writing.
Qianqian Duan: Literature Survey, Application Section Writing, and Editing.
Dawei Zhang: Overall Guidance, Manuscript Structure Optimization, and Resource Coordination.
All authors have read and approved the final version of the manuscript.

\textbf{Conflict of interest} The authors have no relevant financial or non-financial interests to disclose.

\bibliographystyle{plain}

\end{document}